\documentclass[a4paper]{IEEEtran}
\IEEEoverridecommandlockouts

\usepackage{cite}
\usepackage{amsmath,amssymb,amsfonts}
\usepackage[ruled,vlined,linesnumbered]{algorithm2e}
\usepackage{graphicx, subfig, float}
\usepackage{caption}
\usepackage{textcomp}
\usepackage{booktabs}
\usepackage{color,soul}
\usepackage{tabularx,hhline}
\usepackage{multirow}
\usepackage{tikz}
\usepackage{xcolor}
\usepackage{cancel}
\usepackage{enumitem}

\usepackage[strings]{underscore}

\begin{document}
\title{Robust Power System State Estimation using Physics-Informed Neural Networks}

\author{Solon Falas,~\IEEEmembership{Student Member,~IEEE}, Markos Asprou,~\IEEEmembership{Member,~IEEE}, {Charalambos Konstantinou,~\IEEEmembership{Senior Member,~IEEE}, Maria K. Michael,~\IEEEmembership{Member}}
\thanks{This work was partially supported by the European Union’s Horizon 2020 research and innovation programme under grant agreement N0 739551 (KIOS CoE – TEAMING), from the Republic of Cyprus through the Deputy Ministry of Research, Innovation and Digital Policy, the Cyprus Research and Innovation Foundation under grant agreement CODEVELOP-REPowerEU/1223/91 (GridGnosis), through the European Union’s NextGenerationEU initiative and the Recovery and Resilience Plan of Cyprus (Cyprus-Tomorrow), and the University of Cyprus. This publication is partially based upon work supported by King Abdullah University of Science and Technology (KAUST) under Award No. ORFS-CRG11-2022-5021.
}
\thanks{Solon Falas, Markos Asprou and Maria K. Michael are with the Department of Electrical and Computer Engineering, University of Cyprus, KIOS Research and Innovation Center of Excellence, Nicosia, Cyprus
(e-mail: falas.solon@ucy.ac.cy; asprou.markos@ucy.ac.cy; mmichael@ucy.ac.cy)}
\thanks{Charalambos Konstantinou is with the Computer, Electrical and Mathematical Sciences and Engineering (CEMSE) Division, King Abdullah University of Science and Technology (KAUST), Thuwal 23955-6900, Saudi Arabia (e-mail: charalambos.konstantinou@kaust.edu.sa).}}


%



\IEEEaftertitletext{\vspace{-3\baselineskip}}


\maketitle

\begin{abstract}
Modern power systems face significant challenges in state estimation and real-time monitoring, particularly regarding response speed and accuracy under faulty conditions or cyber-attacks. This paper proposes a hybrid approach using physics-informed neural networks (PINNs) to enhance the accuracy and robustness, of power system state estimation. By embedding physical laws into the neural network architecture, PINNs improve estimation accuracy for transmission grid applications under both normal and faulty conditions, while also showing potential in addressing security concerns such as data manipulation attacks. Experimental results show that the proposed approach outperforms traditional machine learning models, achieving up to $\sim$83\% higher accuracy on unseen subsets of the training dataset and $\sim$65\% better performance on entirely new, unrelated datasets.
Experiments also show that during a data manipulation attack against a critical bus in a system, the PINN can be up to $\sim$93\% more accurate than an equivalent neural network.
\end{abstract}

\vspace{-\baselineskip}

\section{Introduction}\label{sec:introduction}
The escalating global electricity demand, driven by rapid urbanization, transportation electrification, and digital technology proliferation, has underscored the need for robust and stable power systems. Ensuring the stability and security of critical infrastructures, particularly power transmission networks, is essential for economic stability and public safety. However, the growing complexity of modern grids, driven by renewable energy integration, adoption of smart grid technologies, and interconnected networks, presents significant challenges in monitoring, control, and system resilience~\cite{alotaibi2020renewables,huang2012state}.

In particular, addressing challenges related to real-time data management and stability has become increasingly critical, necessitating advanced monitoring schemes to ensure system stability and integrity. It is therefore important to have advanced monitoring schemes to ensure the system stability and integrity. Traditional state estimation methods, such as Weighted Least Squares (WLS), rely on Supervisory Control and Data Acquisition (SCADA) systems and iterative processes, but they are limited by insufficient real-time capabilities, high computational demands, and reliance on sparse data~\cite{sanjab2016smartgridsecurity,kosut2010SEattacks}. These limitations are especially problematic with intermittent renewable energy sources like solar and wind, which cause rapid fluctuations and transient errors. Phasor Measurement Units (PMUs) enhance real-time monitoring with high-precision synchronized data, but conventional methods still fall short in addressing the complexity of modern systems~\cite{khanam2022state}.

Compounding these technical challenges, power grids have simultaneously become prime targets for cyber-attacks, such as Denial of Service (DoS) and False Data Injection (FDI)  attacks~\cite{vahidi2023security,liang2016review}. Such sensor manipulation attacks can compromise the integrity of grid monitoring, leading to potential outages, and severe physical damage~\cite{case2016analysis,soltan2018blackiot}. Therefore, a more cyber-physical approach to state estimation is imperative to enhance the resilience of power grids against such threats.

To address these limitations, machine learning (ML) techniques, including neural networks and support vector machines, have been explored to improve state estimation by leveraging statistical patterns and historical data~\cite{weng2016robust}. While these data-driven approaches enhance accuracy and efficiency, they often neglect the physical constraints inherent to power systems~\cite{mestav2019bayesian,zhang2019unrolled}, limiting their reliability under varying operational conditions or sparse data~\cite{markidis2021old}.

PINNs~\cite{raissi2019physics} offer a transformative approach by embedding physical laws, constraints, and boundary conditions directly within the ML framework. By integrating principles such as conservation laws and system constants, PINNs align predictions with the physical realities of power systems~\cite{angulo2024photovoltaic}. This not only improves model accuracy but also reduces data requirements, both in quantity and quality, compared to purely data-driven methods. PINNs' attributes make them particularly suited for applications requiring robustness, accurate fault diagnosis, and rapid responses during disruptions~\cite{lu2024dynamic,misyris2020power}.

This work presents a hybrid framework for power system state estimation that leverages the capabilities of PINNs. The proposed approach seeks to improve both the accuracy and robustness of state estimation, particularly in the presence of disturbances such as cyberattacks and operational anomalies. The primary contributions of this work are:
\begin{enumerate}
\item Proposing a hybrid model that incorporates physical laws into neural networks to improve the accuracy of state estimation in power systems.
\item Conducting a comprehensive benchmark against state-of-the-art data-driven neural networks through an extensive set of experimental evaluations.
\item Demonstrating significant improvements in accuracy (up to $\sim93\%$) and robustness under various challenging conditions, including three-phase faults and data manipulation attacks.
\end{enumerate}
The proposed PINN-based approach highlights the potential for enhancing the stability, integrity, and security of power grids by utilizing advanced ML models that integrate physical principles to enable accurate and timely state estimation.

The paper is organized as follows: Section~\ref{sec:related_work} reviews existing research on physics-enhanced neural networks for state estimation. Section~\ref{sec:methodology} explains the methodology and training process of the proposed model. Section~\ref{sec:exp_results} presents experimental results, and Section~\ref{sec:conclusion} summarizes the study's key insights and future directions.

\section{Related Work}\label{sec:related_work}
State estimation methods in power systems can be broadly categorized into three main categories: traditional techniques, data-driven techniques, and hybrid models~\cite{Dehghanpour2019surveyestimation,zhao2019power}. Traditional techniques, such as Newton-Raphson and Weighted Least Squares, use physical laws to provide reliable results under steady-state conditions. However, these methods are computationally expensive and show limited ability to handle dynamic conditions. Data-driven techniques, particularly Neural Networks, offer advantages in handling complex nonlinearities in historical data, with faster execution than traditional methods. Yet, these approaches may lack physical consistency and are sensitive to overfitting, requiring large datasets for reliable performance and accuracy. Hybrid models, such as PINNs, bridge this gap by utilizing the strengths of data-driven approaches and combining them with physical properties, while maintaining the ability to handle data scarcity.

Machine learning techniques for state estimation have advanced beyond traditional methodologies by integrating historical data and statistical patterns~\cite{mestav2019bayesian,zhang2019unrolled}. While data-driven approaches offer compelling advantages, they frequently overlook fundamental physical constraints~\cite{markidis2021old}, potentially compromising estimation accuracy under data-limited or dynamically varying scenarios. Hybrid modeling frameworks, synthesizing data analytics with governing physical principles~\cite{huang2022applications} emerge as a promising paradigm, enabling near real-time system state characterization with enhanced robustness in cyber-physical system applications.

PINNs augment neural network architectures by systematically incorporating physical constraints as regularization terms during model training~\cite{raissi2019physics}. This innovative approach fundamentally transforms machine learning methodologies by thoughtfully integrating domain-specific physical knowledge into the learning process. By embedding first-principles constraints, PINNs generate physically consistent predictions with reduced data requirements compared to conventional neural network approaches. Their nuanced versatility in managing data-scarce scenarios, processing multidimensional inputs, and preserving physical system integrity renders them particularly promising for sophisticated power system modeling~\cite{falas2023physics,misyris2020power}.

On the other hand, due to the digitization of power systems applications, many functionalities of energy management systems are  increasingly susceptible to cyberattacks, such as data manipulation and false data injection attacks. Such attacks often target state estimation routines and can lead to severe operational failures, compromising grid security~\cite{tatipatri2024comprehensive}. Purely data-driven state estimators are often vulnerable to these manipulations. In contrast, hybrid methodologies enhance resilience by embedding physical laws (e.g., Kirchhoff's laws) into the data-driven framework~\cite{rahman2024adversarial}. This enables the validation of measurement data against underlying physical principles, offering a pathway to protect systems against attacks that violate these constraints~\cite{gaggero2025artificial}.

Recent research integrating physical constraints into neural networks for power system state estimation, while promising, faces limitations. A $\mu$PMU-based partitioning strategy is proposed in~\cite{zamzam2020distribution}, and a deep neural network model focused on limited observability distribution grids in~\cite{ostrometzky2020pinn}. Moreover, a graph neural network using Kron reduction, which addresses transmission grids~\cite{pagnier2021physics}, demonstrated parameter estimation accuracy improvements. However, these methods share common limitations. Notably, they lack rigorous regularization parameter selection, which hinders application across diverse grid configurations and systematic model optimization. Furthermore, their evaluation primarily considers typical load profiles, neglecting critical outlier scenarios such as attacks and faults, essential for grid resilience. These events, while infrequent, can severely impact grid stability and lead to failures.

Physics-informed models for power system state estimation have utilized unrolled iterative methods for acceleration~\cite{yang2022data,pagnier2021embedding}. While promising and robust to input perturbations, their resilience to diverse data corruption, including faults and unexpected behaviors, is limited. Additionally, time-series models may exhibit biases from time-series specific periodicities. The robustness of PINNs against critical system anomalies and diverse operating conditions remains largely unexplored in power systems~\cite{lu2024dynamic,andronikidis2024anomaly}. This work addresses these limitations by proposing a methodology designed for enhanced robustness against diverse operating conditions, such as faults and cyber threats. Our approach is time-series agnostic, offers a systematic hyperparameter optimization procedure, and embeds physics laws directly into the loss function, aiming for high estimation accuracy and consistency under challenging conditions.
\section{Methodology}\label{sec:methodology}
\subsection{Background}\label{ssec:background}
The foundational concept behind PINNs~\cite{raissi2019physics} is to incorporate physical laws directly into the training process of neural networks by embedding these laws as constraints in the loss function. In general, PINNs can approximate a target function \( f \) by training a neural network \( \hat{f}(\mathbf{z}; \theta) \), parameterized by weights \( \theta \), where \( \mathbf{z} \) represents input variables. Instead of relying purely on data, PINNs include physics-based constraints that govern the system's behavior, ensuring that the model output conforms to known physical laws.

The PINN loss function combines a data-driven term, \( L_{\text{data}} \), which minimizes prediction error on observed data, and a physics-informed term, \( L_{\text{physics}} \), which enforces consistency with the underlying physical laws:

\begin{equation}
L_{\text{data}} = \sum_{i=1}^N \| \hat{f}(\mathbf{z}_i; \theta) - f_i \|^2,
\end{equation}

where \( \{(\mathbf{z}_i, f_i)\}_{i=1}^N \) represents observed data points, and

\begin{equation}
L_{\text{physics}} = \sum_{j=1}^M \| G(\hat{f}(\mathbf{z}_j; \theta), \mathbf{z}_j) \|^2.
\end{equation}

Here, \( G(\hat{f}, \mathbf{z}) = 0 \) denotes a general physical law or system constraint that should hold at each point \( \mathbf{z}_j \) in the domain, represented by \( \{ \mathbf{z}_j \}_{j=1}^M \).

The combined loss is expressed as:

\begin{equation}
L = L_{\text{data}} + \lambda L_{\text{physics}},
\end{equation}

where \( \lambda \) is a weighting factor that balances the data-driven and physics-informed components of the loss.

During training, minimizing \( L \) guides the model \( \hat{f} \) to align with observed data while satisfying the physical constraints, allowing PINNs to model complex behaviors accurately and consistently with the underlying physical system. This approach enhances model generalization and reduces dependence on extensive data by leveraging embedded physics, making PINNs particularly effective for applications in scientific and engineering domains.

\begin{figure*}[t]
    \centering
    \includegraphics[width=\textwidth]{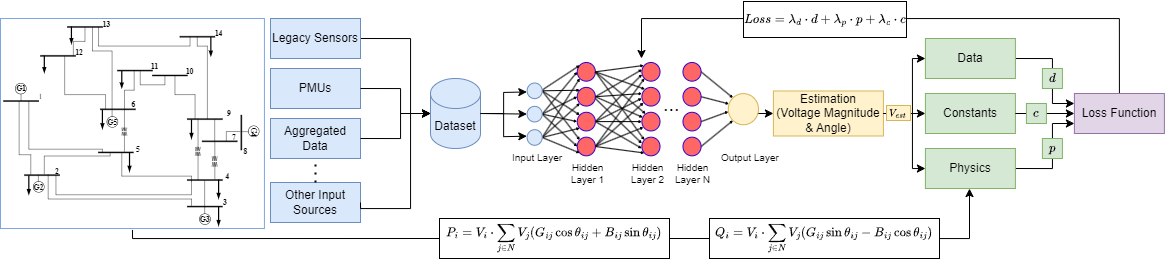}
    \caption{Model training procedure for the power system state estimation use case.}
    \label{fig:training_diagram}
    \vspace{-\baselineskip}
\end{figure*}

\subsection{Enhancing Loss Functions with Physical Constraints}\label{ssec:loss_function}
In this study, we build upon the foundational principles of PINNs by systematically introducing additional terms to the loss function that impose physics-based constraints relevant to power system state estimation. These terms are derived from established relationships and laws governing the system, such as power flow equations, Kirchhoff's laws, and the admittance matrix. By embedding these constraints directly into the loss function, we ensure that the neural network's predictions align with the physical laws governing power systems. Consequently, during the training process, the network aims to minimize not only the discrepancy between predicted and actual values but also the violation of these physics-based constraints. This approach enhances the model's accuracy and reliability, making it particularly suitable for the complexities inherent in power system state estimation.

In the proposed PINN method, the loss function (denoted as $Loss$) is computed as the sum of the \textit{Mean Square Error (MSE)} between the actual data and the inferred values ($d$) and the \textit{MSE} resulting from enforcing a physics equation ($p$) that should ideally equal zero. Also, if certain constants, such as the admittance matrix or specific system parameters, are known and remain unchanged ($c$), they can be included throughout the training process. Let $\lambda_{d},~\lambda_{p},~\lambda_{c}~\in~[0,1]$ denote the weights for data, physics, and constants, respectively. These weights can be adjusted during the training process to optimize the model's performance based on the influence of each loss function parameter. Clearly, $\lambda_{d}~+~\lambda_{p}~+~\lambda_{c}~=~1$.
The $Loss$ function can be generally represented as:
\begin{equation}\label{eq:loss_function}
Loss = \lambda_{d} \cdot d + \lambda_{p} \cdot p + \lambda_{c} \cdot c
\end{equation}
By integrating physics-based constraints as regularization terms, along with adjustable weighting factors, we can achieve a balance between data-driven learning and adherence to fundamental physical laws and properties. This is also helpful in avoiding overfitting of the neural network to the training data while penalizing physically unrealistic results.

\subsection{Power System Parameter Selection}\label{ssec:parameter_selection}
To develop the proposed PINN model, an input-output pair is first identified, where the input is correlated to the output through a physics-defined relationship that captures the system's dynamic nature. It is particularly beneficial if this relationship includes constants or topological information about the system, as this introduces additional information and constraints to the model. Next, the loss function is formulated to incorporate data, physics-based constraints, and constants, ensuring these factors are considered during the model's training process. During parameter optimization, the weights assigned to each component of the hybrid loss function shown in Eq.~\ref{eq:loss_function} are adjusted according to the specific characteristics of the system.

The aforementioned procedure for determining the PINN model is  facilitated through the consideration of a traditional power system state estimation. In particular, the process of power system state estimation involves determining the voltage magnitude ($V_i$) and phase angle ($\theta_i$) at each bus within the system. In the proposed PINN model, the outputs are the estimated complex voltages, while the inputs are the active power ($P_i$) and reactive power ($Q_i$) injection measurements. These measurements are chosen as inputs because they play a critical role in revealing the system's topology through data-driven patterns. The active and reactive power injections are calculated using the following equations:
\begin{subequations}\label{eq:power_flow}
\begin{equation}
    P_i = V_i \cdot \sum_{j \in N} V_j (G_{ij}\cos{\theta_{ij}} + B_{ij}\sin{\theta_{ij}})
\end{equation}
\begin{equation}
    Q_i = V_i \cdot \sum_{j \in N} V_j (G_{ij}\sin{\theta_{ij}} - B_{ij}\cos{\theta_{ij}})
\end{equation}
\end{subequations}
where $V_i$ and $V_j$ represent the voltage magnitudes at buses $i$ and $j$, respectively, $\theta_{ij}$ denotes the phase angle difference between buses $i$ and $j$, $G_{ij}$ and $B_{ij}$ are the real and imaginary components of the admittance matrix, and $N$ is the total number of buses in the system. As indicated by equations (2a) and (2b), the net power injected at a bus is determined by the complex voltage at that bus and the voltages of the buses connected to it.

The power system's topology and connectivity are represented by the admittance matrix included in these equations ($G_{ij}$ and $B_{ij}$). By learning the correlations between complex powers and voltages, the neural network can uncover the underlying system topology without the need to explicitly modify the neural network architecture with system-specific parameters or special connectivity arrangements between the network's layer.

\subsection{Loss Function Formulation}\label{ssec:loss_formulation}
The loss function for the PINN, as formulated in Eq.~\ref{eq:loss_function}, is designed to be minimized during the backpropagation training process, as depicted in Fig.~\ref{fig:training_diagram}. The specific parameters employed in the loss function are derived from the input-output relationships detailed in Section~\ref{ssec:parameter_selection}. Minimizing this loss function enables the PINN to produce outputs that closely approximate the desired power system state. Backpropagation, a widely used technique in neural network training, iteratively updates the network's weights and biases to reduce the loss by propagating the error backward through the network. This optimization process aims to identify the parameter configuration that minimizes the error between estimated and true values, thus enhancing the performance of the power system state estimation.

Based on Eq.~\ref{eq:loss_function}, parameter $d$, a very standard loss function parameter aims to drive the model towards matching the results of a known, ground-truth dataset. The model produces a pair of voltage magnitude $V_{mag}$ and voltage angle $V_{ang}$ for each bus at the output layer, constituting the complex voltage $V_{est}$. In order to ensure that the model fits the data according to a known, representative dataset, the difference between the estimated and ground-truth values $V_{true}$ should be minimized. Thus, the $d$ parameter is constructed as the mean squared error between the estimated and the actual complex voltages, as shown below:
\begin{equation}
    d = \frac{1}{n} \sum_{i=1}^{n} (V_{{est}_i} - V_{{true}_{i}})^2
\end{equation}

For the physics-determined part of the loss function $p$, the complex currents injected at each bus are used. Actually, the complex currents relate to the system admittance matrix $\textbf{Y}$ (that includes information about the network connectivity) with the complex voltages as shown below:
\begin{equation}\label{eq:complex_impedance}
    \textbf{I} = \overline{\textbf{Y}} \cdot \textbf{V}
\end{equation}
Through Eq.~\ref{eq:complex_impedance}, the estimated current $I_{est}$ is calculated using the complex voltage values generated at the output layer of the model. Furthermore, Eq.~\ref{eq:complex_impedance} is used to calculate the actual injected current using the ground-truth data from a given dataset. Having these datasets available, the mean squared error between the actual and the estimated injected current is calculated and attributed in $p$ as shown in Eq.~\ref{eq:loss_mse_physics}.
\begin{subequations}\label{eq:loss_physics_part}
    \begin{equation}
        I_{est} = \overline{\textbf{Y}} \cdot V_{est}
    \end{equation}
    \begin{equation}
        I_{true} = \overline{\textbf{Y}} \cdot V_{true}
    \end{equation}
    \begin{equation}\label{eq:loss_mse_physics}
        p = \frac{1}{n} \sum_{i=1}^{n} (I_{{est}_i} - I_{{true}_{i}})^2
    \end{equation}
\end{subequations}
The use of the mean squared error of the injected currents ensures that both $V_{est}$ and $I_{est}$ are considered during the training procedure. As the current is correlated to the voltage through the admittance matrix, during the training procedure the PINN model is pushed towards adhering to the system's topology.

The constants of the system, represented by $c$, are encoded in the loss function as a mean squared error between a known value of the system and the model's predicted value. For example, in a power system, there are buses connected to generators where a voltage regulation procedure is carried out. Therefore, we know that the voltage magnitude or angle in certain buses is going to be considered a constant, specifically during steady state conditions. Also, in the case of steady state conditions, it can be assumed that the slack bus is designated as a reference point and the associated voltage and angle will be constant as well.
Therefore, $c$ can be calculated as the sum of mean squared errors between known constants and the predicted values, so that we can incentivize the model to fit these known values:
\begin{equation}
    c = \frac{1}{n} \sum_{i=1}^{n} (C_{{est}_i} - C_{{true}_{i}})^2
\end{equation}

The methodology and criteria outlined in this section, systematically guide the selection of measurements that constitute the loss function, ensuring adaptability to various scenarios. As such, the parameters chosen can be substituted with alternative equations and system measurements, depending on the specific application of the PINN. In our experimental validation, Eq.~\ref{eq:loss_function} will be employed, with the parameters $d$, $p$, and $c$ defined as discussed in this section. These equations have been specifically tailored for the exploration of a power system state estimation use case.

\subsection{Systematic Hyperparameter Optimization}\label{ssec:hyperparameters_&_dataset}

Hyperparameter optimization is crucial for training complex neural network architectures, such as the proposed PINN model. Given the inefficiency of traditional grid and random search methods in exploring the often vast hyperparameter space, especially when balancing multiple loss function components and model parameters, a systematic approach using meta-optimizers is employed. This approach, outlined in Algorithm~\ref{alg:training_process}, aims to determine the optimal configuration of adjustable loss function weights for each scenario.

\SetAlgoSkip{0pt} 

\newlength{\originalbelowcaptionskip} 
\setlength{\originalbelowcaptionskip}{\belowcaptionskip}
\setlength{\belowcaptionskip}{0em}

\begin{algorithm}[t]
    \SetAlgoLined
    \SetKwInOut{Input}{Input}
    \SetKwInOut{Output}{Output}
    \SetKwInOut{Parameter}{Parameters}
    
    \caption{Model Optimization Process}
    \label{alg:training_process}
    \Input{
        $k$:~number of weights, \
        $n$:~step size, \
        $t$:~number of trials, \newline
        $param\_ranges$: (layers, number of neurons, learning rate, batch size), \newline
        $pinn$: tuple $pinn(m,E)$ denotes model $m$ with testing Mean Absolute Error (MAE), $E$, of model $m$
    }
    \Output{
        $pinn_{opt}$: Outputs best model based on lowest MAE
    }
    
    \tcp{Initialization step}
    $N \gets \binom{n + k - 1}{k - 1}$; \\
    $pinn_{opt} \gets (\text{none},\infty)$; \\
    \tcp{Optimization Procedure}
    \For{$i \gets 1$ \KwTo $N$}{
        $pinn_{trial} \gets (\text{none},\infty)$; \\
        \tcp{Select loss function weight combination for iteration $i$}
        $l\_weights_{i} \gets \text{combination}_{i}(n,k)$; \\
        \For{$j \gets 1$ \KwTo $t$}{
            \tcp{Select trial hyperparameters using the TPE meta-optimizer}
            $params_{ij} \gets \text{TPE}(param\_ranges)$; \\
            $pinn_{ij} \gets \text{model\_train}(params_{ij}, l\_weights_{i})$; \\
            \If{$pinn_{trial}.E > pinn_{ij}.E$}{
                $pinn_{trial} \gets pinn_{ij}$;
            }
        }
        \If{$pinn_{opt}.E > pinn_{trial}.E$}{
            $pinn_{opt} \gets pinn_{trial}$;
        }
    }
    \KwRet{$pinn_{opt}$}
\end{algorithm}

\setlength{\belowcaptionskip}{\originalbelowcaptionskip}

The exploration process systematically investigates combinations of loss function weights $\lambda_{d},\lambda_{p},\lambda_{c}\in[0,1]$, which represent the relative importance of the loss function terms. These weights are constrained to the range $[0,1]$ and are sampled at intervals defined by a step size of $\Delta$. Additionally, the weights must satisfy the condition $\lambda_d + \lambda_p + \lambda_c = 1$, introducing a constraint on their possible combinations. The process systematically evaluates all possible combinations of these weights, where the total number of combinations $N$ is computed using the formula for combinations with repetition:
\[
N = \binom{n + k - 1}{k - 1}, \quad \text{where } n = \frac{1}{\Delta}, \text{ and } k = 3.
\]
where $n$ represents the number of discrete values for each weight, and $k = 3$ corresponds to the number of weights.

For each weight combination, multiple optimization trials $t$ are performed. During each trial, a meta-optimizer, a Tree-Structured Parzen Estimator (TPE) in our case, selects the model's architecture and training parameters-such as the number of layers, neurons per layer, learning rate, and batch size—from predefined ranges. These trials aim to identify configurations that best align with the current weight combination, enabling a thorough exploration of the parameter space.

The optimization process systematically explores critical model parameters: layer count determines the model's representational capacity, neuron count impacts pattern recognition, learning rate controls gradient descent step size, and batch size influences computational efficiency and generalization. By employing MAE as the objective function, the approach enables a comprehensive evaluation of model performance, allowing for adaptive exploration of the parameter space.
\section{Experimental Setup \& Results}\label{sec:exp_results}
\subsection{Experimental Setup}\label{ssec:exp_setup}
To evaluate the effectiveness of our proposed PINN model for power system state estimation, we conducted extensive simulations using the IEEE 14-bus and the IEEE 118-bus benchmark systems and generated realistic power system scenarios using the PowerWorld software~\cite{peyghami2019standard}. The PINN model is implemented in Python with TensorFlow for efficient neural network training and trained on a variety of measurements collected under different operating conditions and disturbances. In this context, a fixed transmission grid topology with a known admittance matrix (Y) is assumed, where each bus provides real and reactive power injections. Under normal conditions, full system observability is maintained, and buses with generators are voltage-regulated. However, during transients, voltage regulation may be compromised, and no parameter is assumed constant.

The models' generalization capabilities and its ability to handle real-world uncertainties are enhanced by introducing simulated measurement noise into our datasets~\cite{asprou2013effect}. This noise injection methodology systematically emulates the inherent measurement inaccuracies and stochastic fluctuations characteristic of complex power system environments. By deliberately incorporating controlled noise patterns, the models' capacity to handle unpredictable signal variations and measurement imprecision is significantly improved. Data preprocessing also included standardization, min-max scaling, and replacing zero values with a small epsilon ($1 \cdot e^{-8}$). Inputs such as real ($P_i$) and reactive ($Q_i$) power are standardized using their mean and standard deviation, then rescaled to $[-1, 1]$ to align with the $\tanh$ activation function.

Neural network weights are initialized with the popular Glorot scheme~\cite{glorot2010understanding}, biases are set to zero, and training utilizes an early stopping strategy with a 20-epoch patience threshold, capping optimization at 100 epochs. The Adam optimizer and $\tanh$ activation function are employed, with the best-performing model (minimizing MAE) selected to prevent overfitting and preserve generalization performance. The models are optimized using the procedure shown in Alg.~\ref{alg:training_process} using the following ranges for key parameters: number of layers (2-10), neurons per layer (64-4096), learning rate ($1 \cdot e^{-5}$ to $1 \cdot e^{-1}$), and batch size (4–128). Also, the number of weights is  set to $k=3$, the step size $\Delta$=0.1 and number of trials $t=10$. The models' computational costs are measured during multiple optimization trials for training and during testing for inference. The average computational cost is outlined in Table \ref{tab:computational_costs}. All measurements are derived from a single NVIDIA RTX 4000 Ada Generation GPU.

\begin{figure}[t]
    \centering
    \subfloat[MAE per training epoch\label{fig:steady_state_mae_per_epoch}]{
        \includegraphics[width=0.30\textwidth]{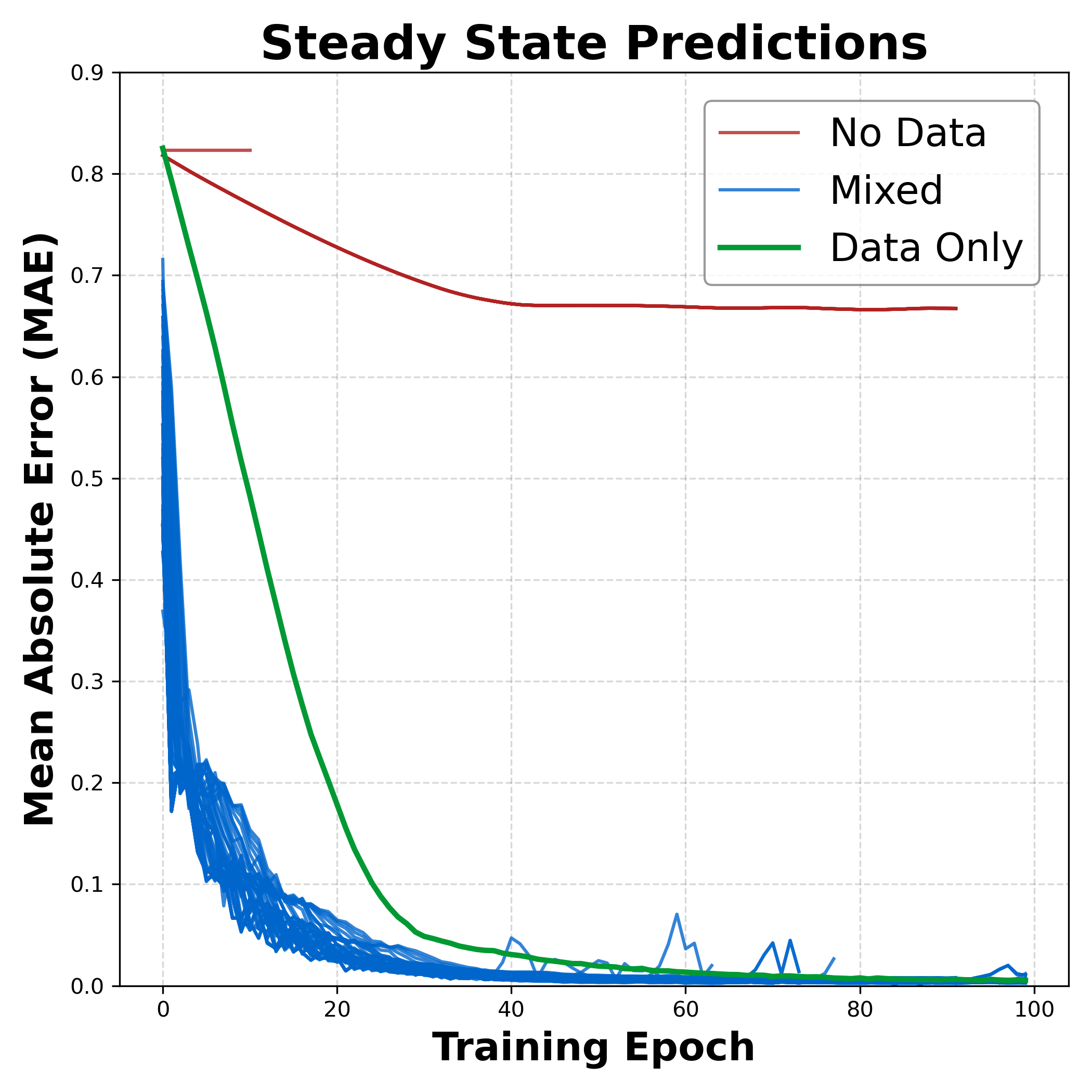}
    }
    \hfill
    \subfloat[$\lambda$ weights heatmap\label{fig:weights_heatmap}]{
        \includegraphics[width=0.40\textwidth]{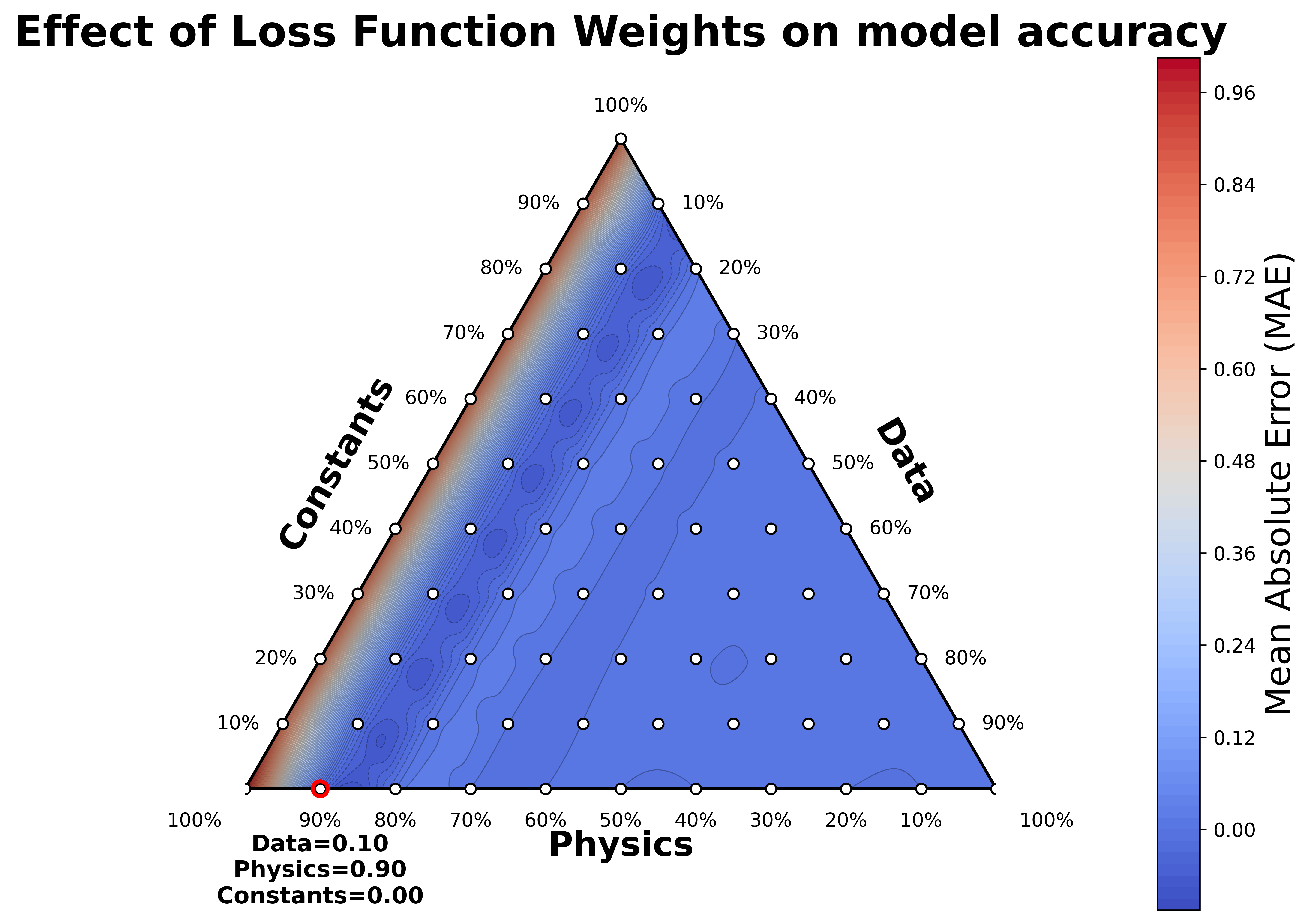}
    }
    \caption{\textit{Scenario 1 (IEEE 14-bus System)} - MAE for steady-state conditions using varying $\lambda_{d},\lambda_{p},\lambda_{c}$ weights for data, physics, and constants, respectively, in the loss function.}
    \label{fig:steady_state}
    \vspace{-\baselineskip}
\end{figure}

\begin{figure*}[t]
    \centering
    \includegraphics[width=0.02\textwidth]{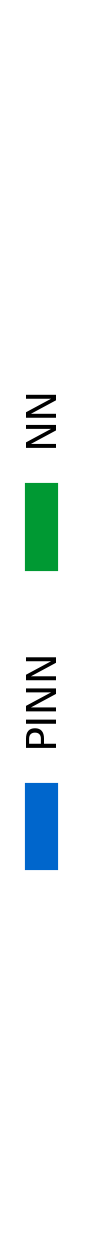}
    \includegraphics[width=0.30\textwidth]{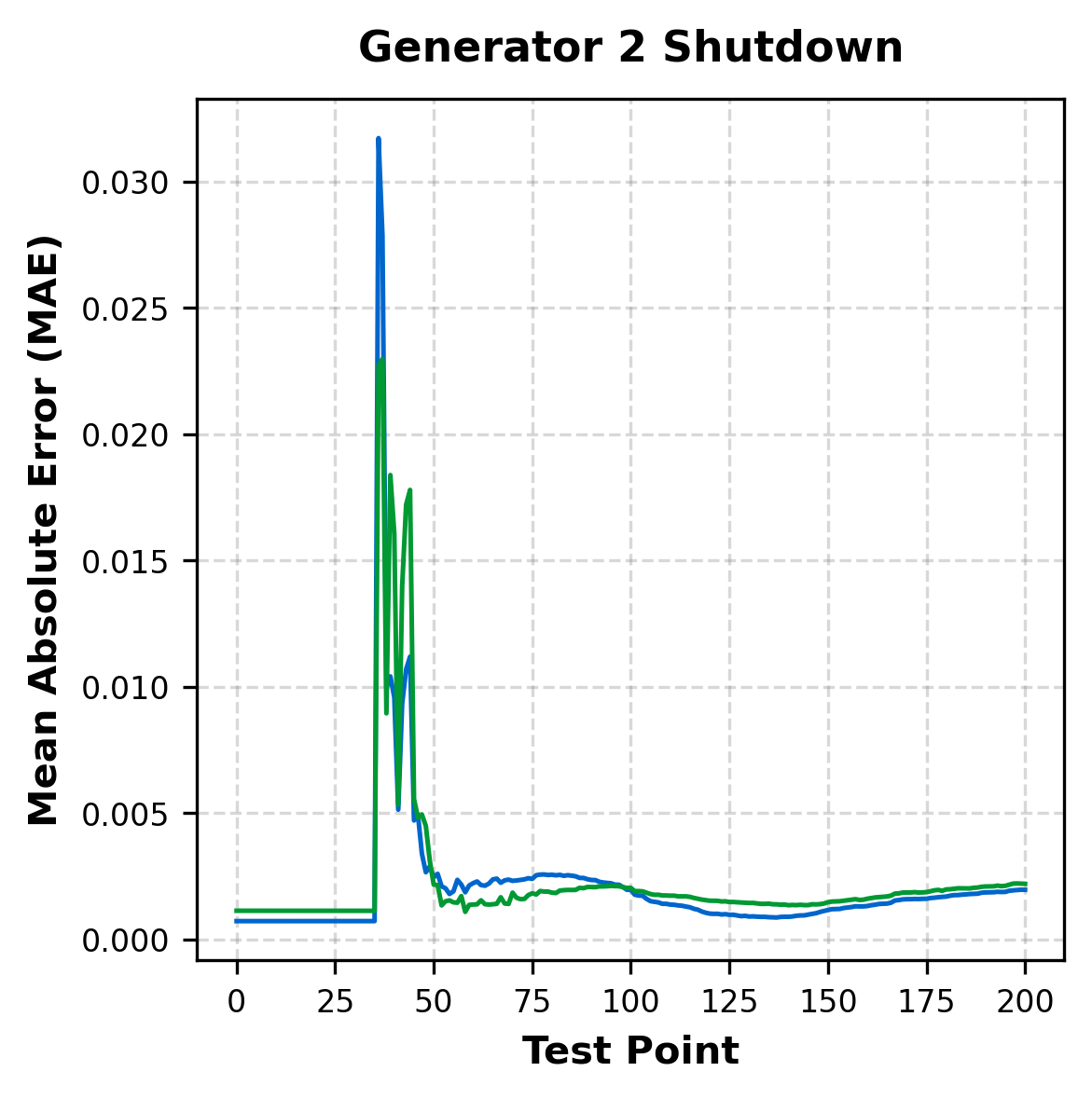}
    \includegraphics[width=0.295\textwidth]{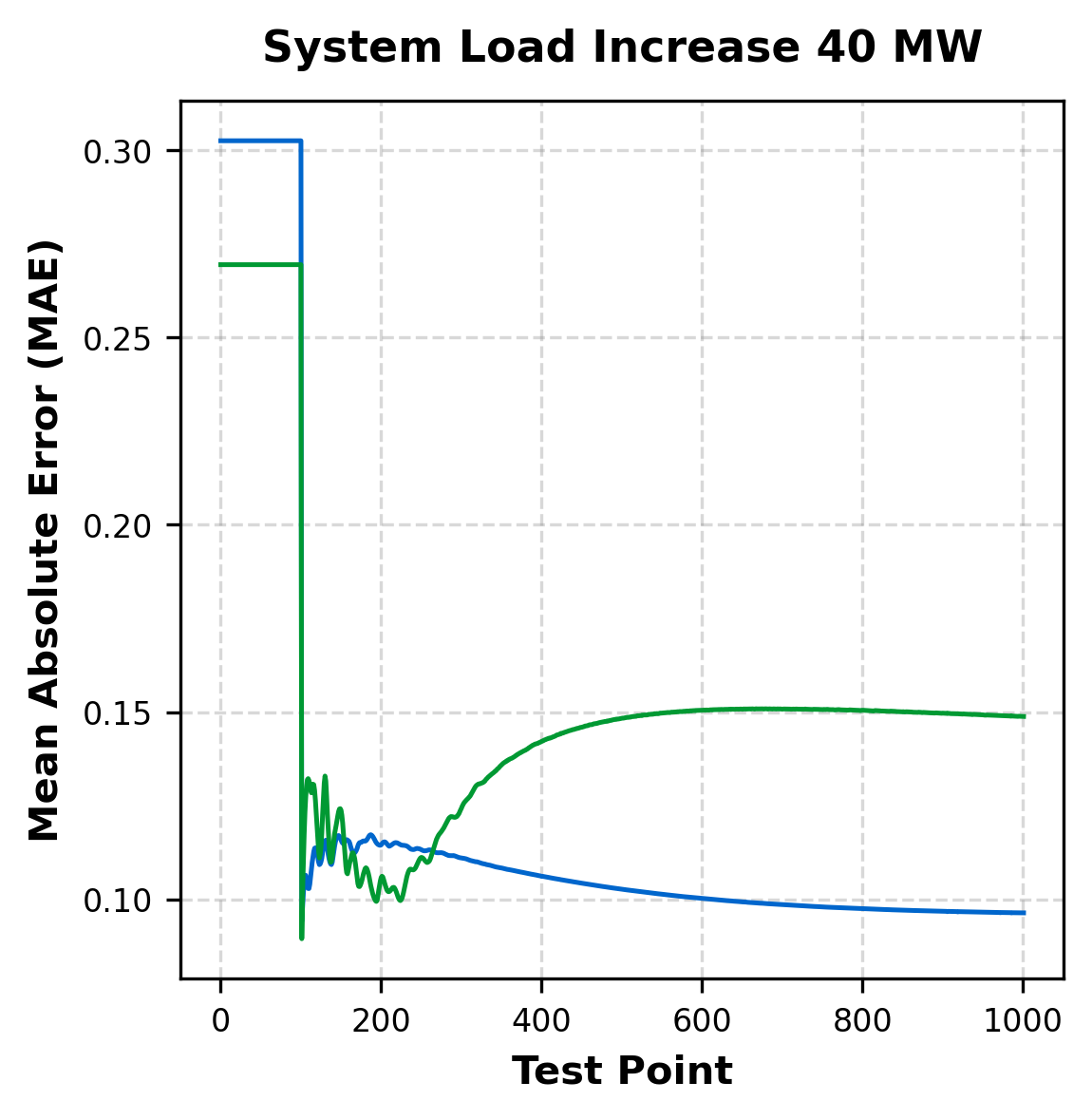}
    \includegraphics[width=0.295\textwidth]{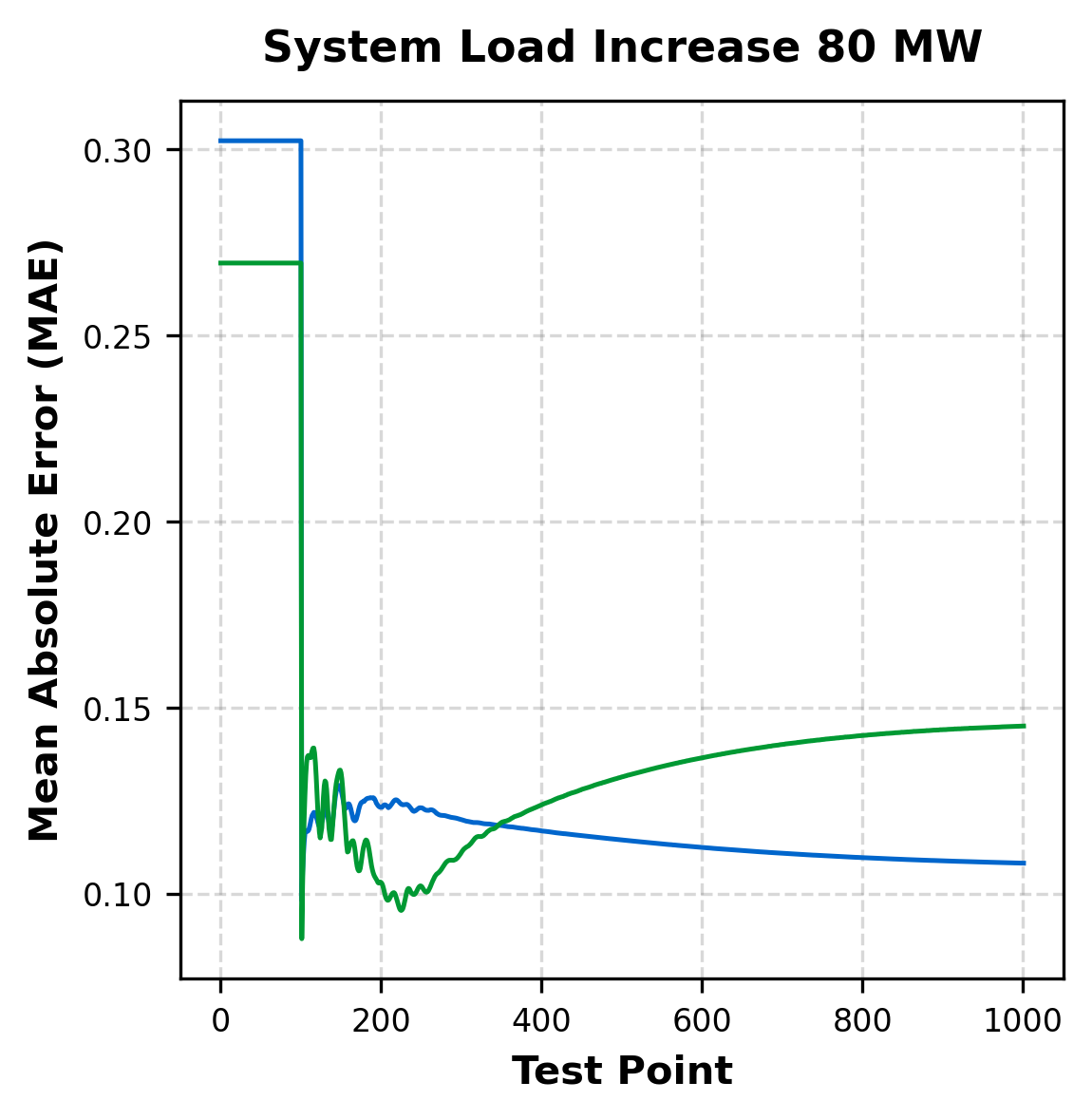}
    \caption{\textit{Scenario 2} - Testing datasets with varying generator shutdowns. Models trained on a bus 2 shutdown are  tested with load spikes at equal (\textit{S2.1}) and double the initial shutdown magnitude (\textit{S2.2}).}
    \label{fig:shutdown}
    \vspace{-\baselineskip}
\end{figure*}

\begin{figure*}[t]
   \centering
   \includegraphics[width=0.02\textwidth]{figures/labels_tilted.png}
   \includegraphics[width=0.295\textwidth]{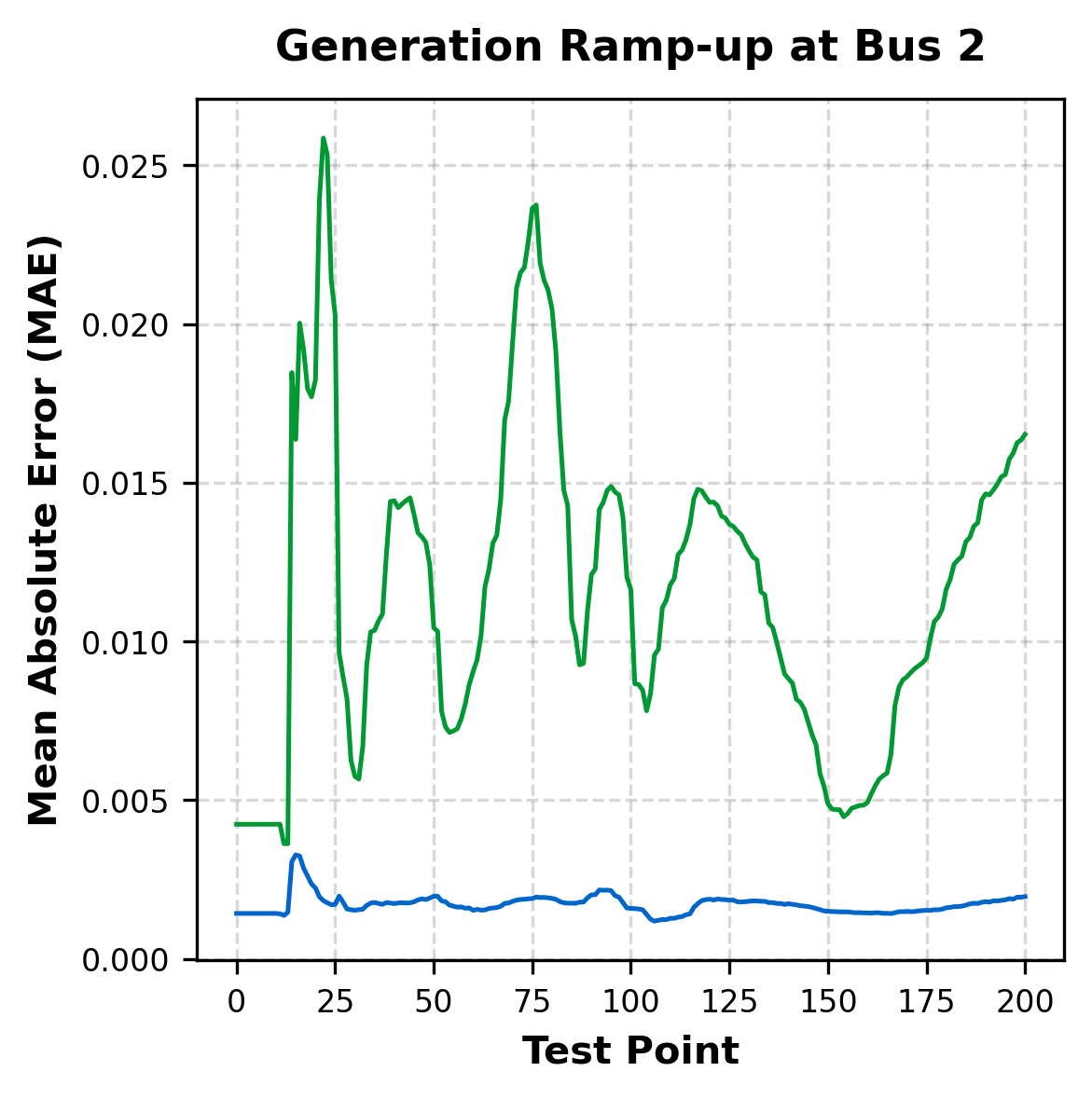}
   \includegraphics[width=0.295\textwidth]{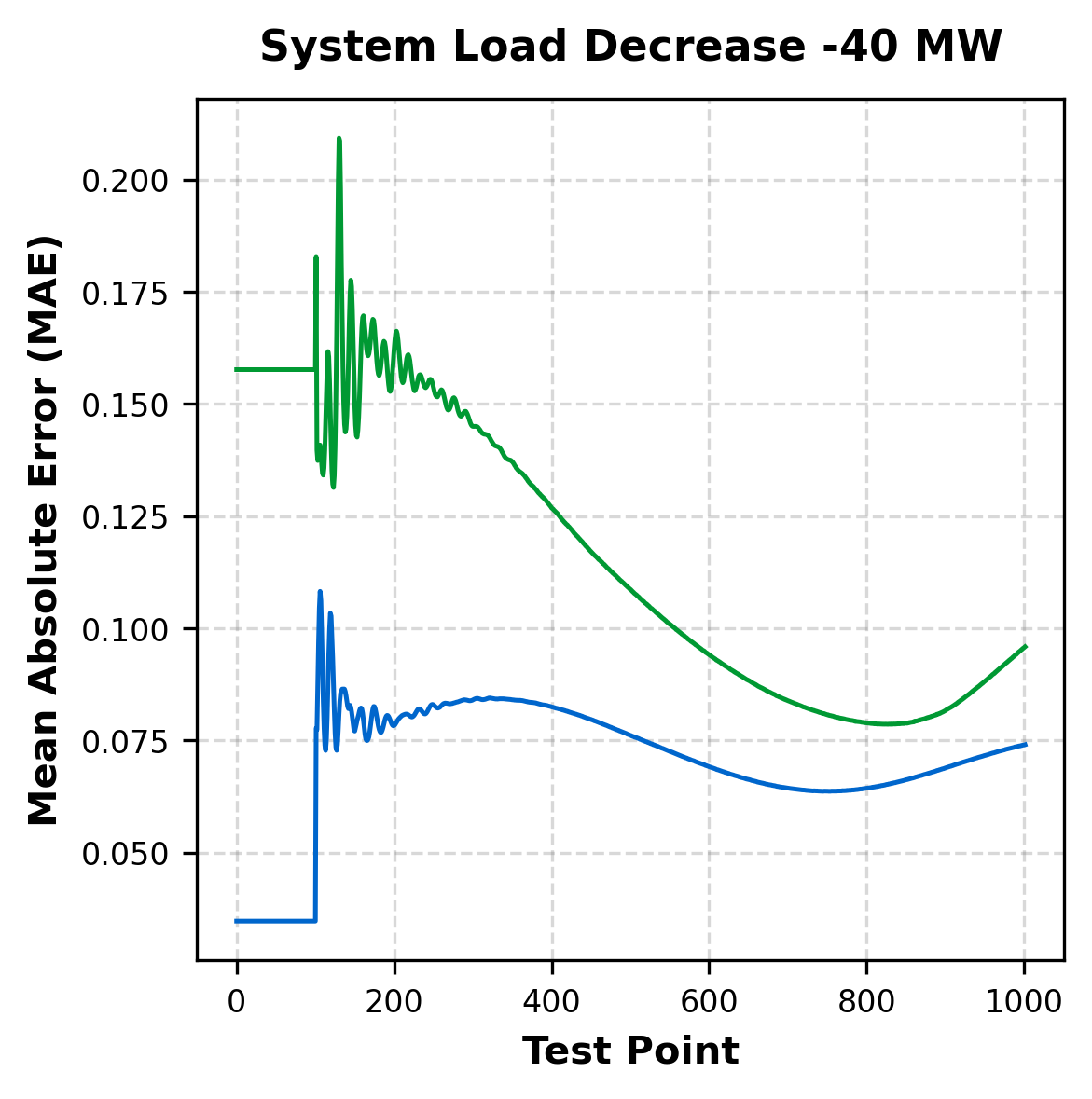}
   \includegraphics[width=0.295\textwidth]{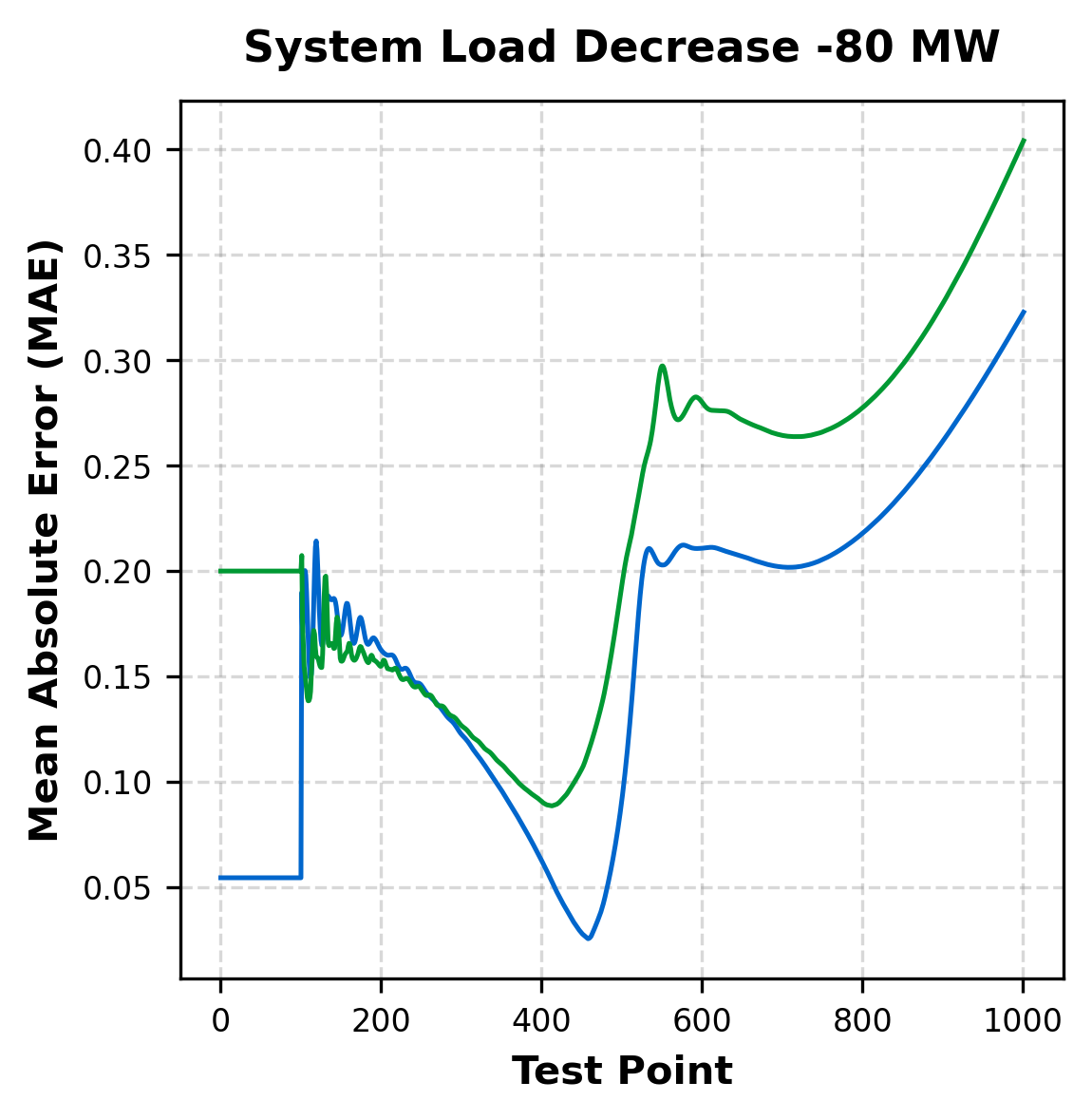}
   \caption{\textit{Scenario 3} - Testing datasets with varying generation increases. Models trained on a bus 2 ramp-up are  tested with scenarios simulating load drops at equal (\textit{S3.1}) and double the initial increase (\textit{S3.2}).}
   \label{fig:increase}
   \vspace{-2\baselineskip}
\end{figure*}

\begin{figure*}[t]
   \centering
   \includegraphics[width=0.0175\textwidth]{figures/labels_tilted.png}
   \includegraphics[width=0.235\textwidth]{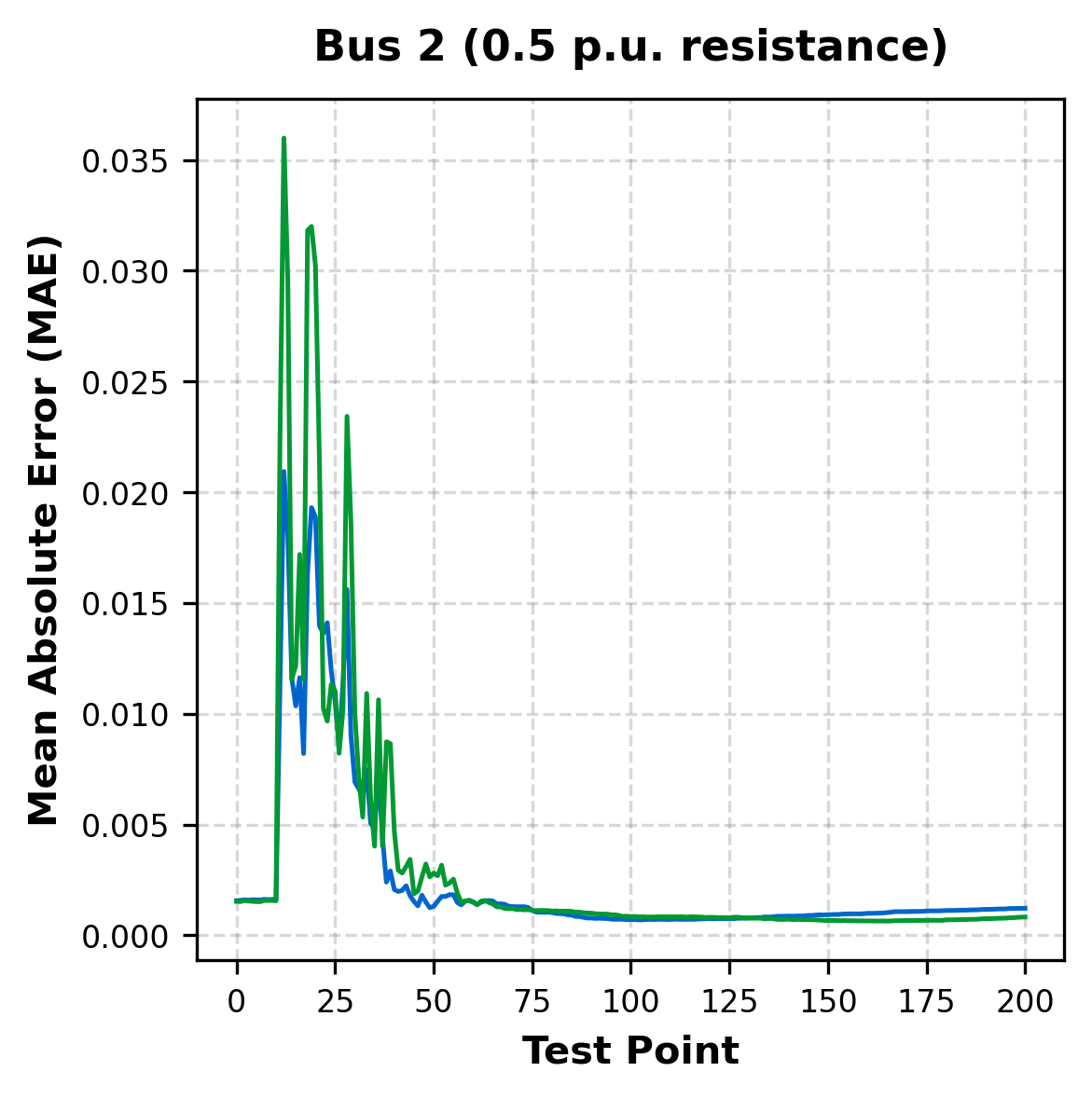}
   \includegraphics[width=0.235\textwidth]{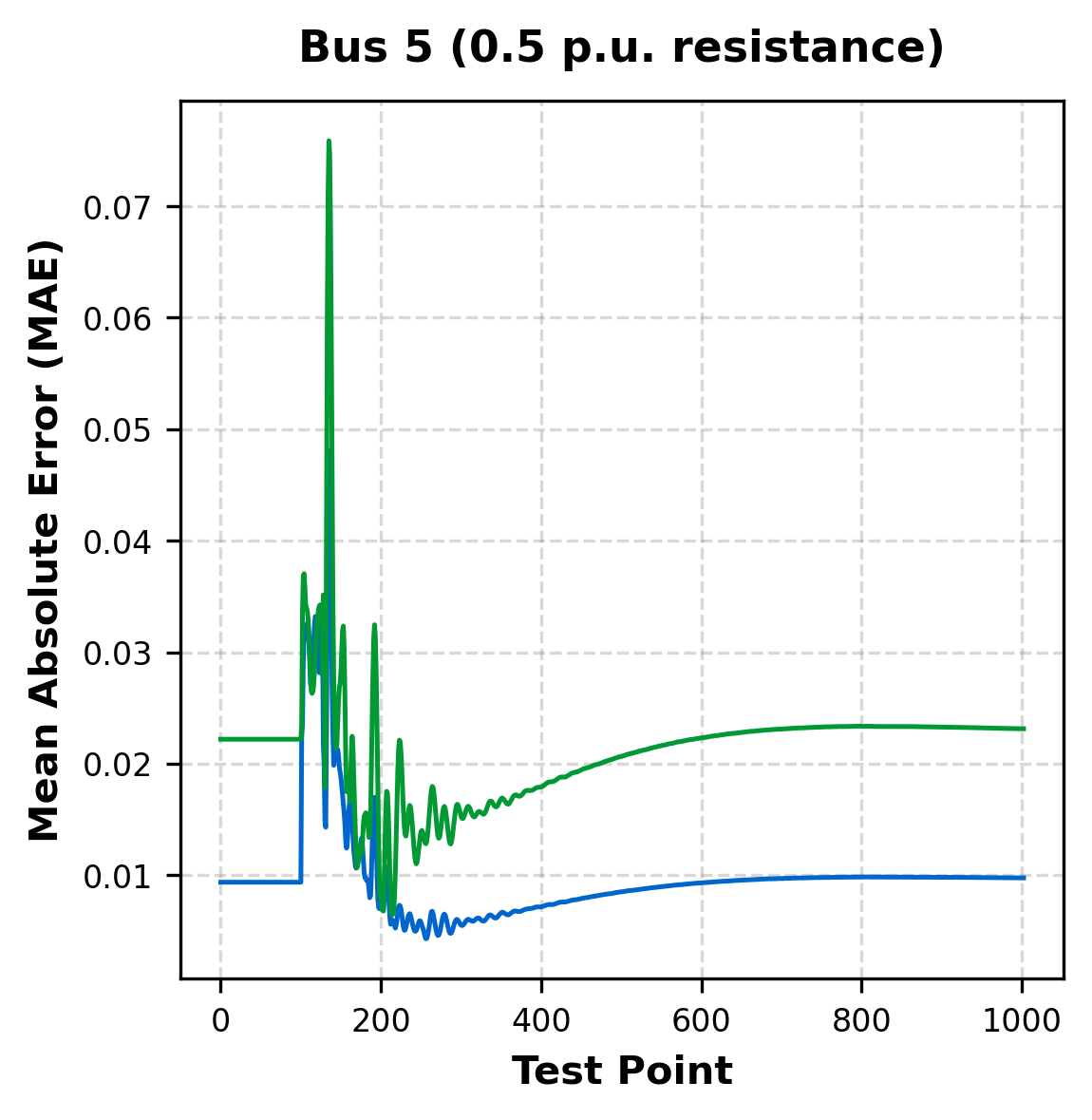}
   \includegraphics[width=0.235\textwidth]{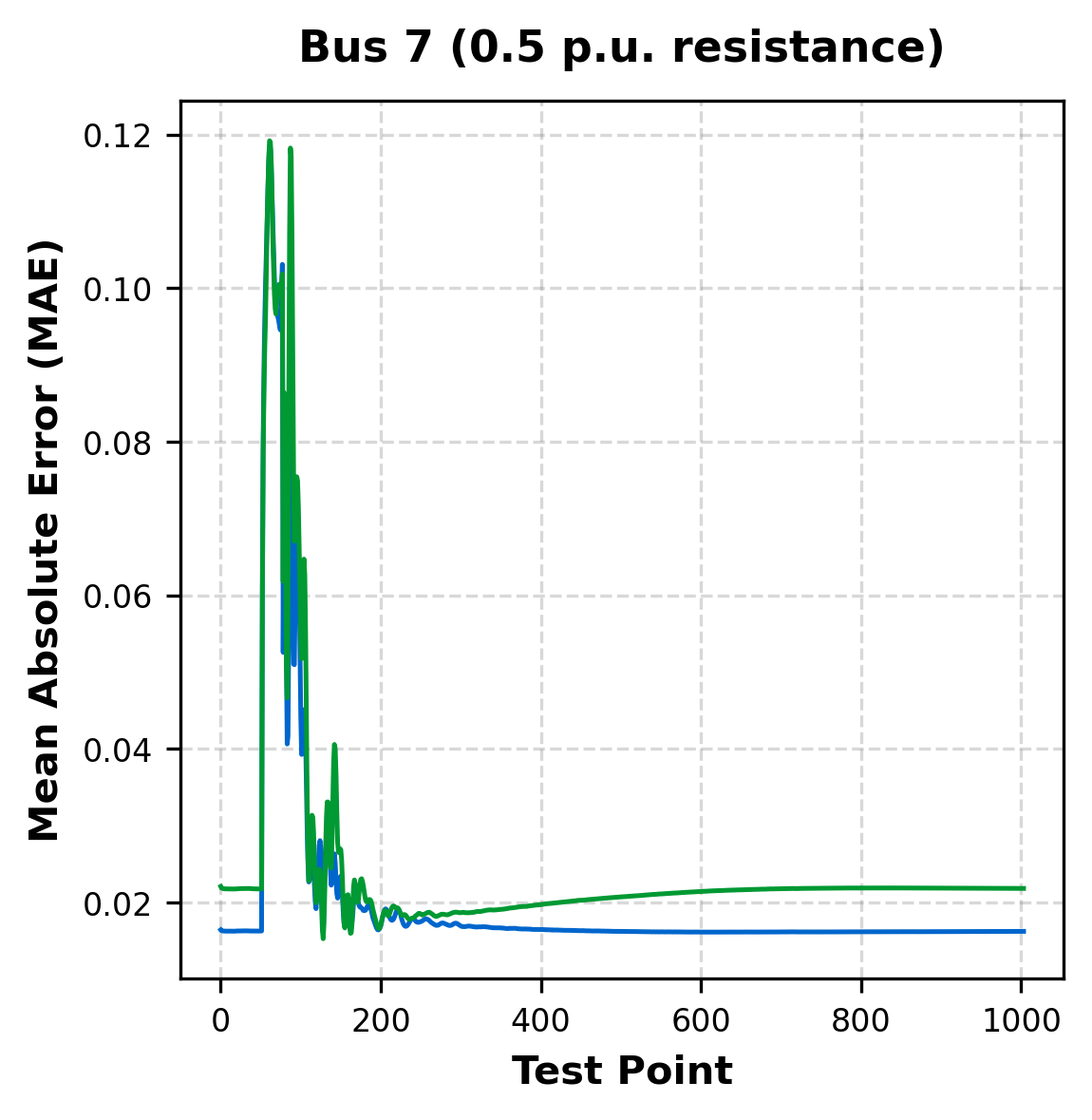}
   \includegraphics[width=0.235\textwidth]{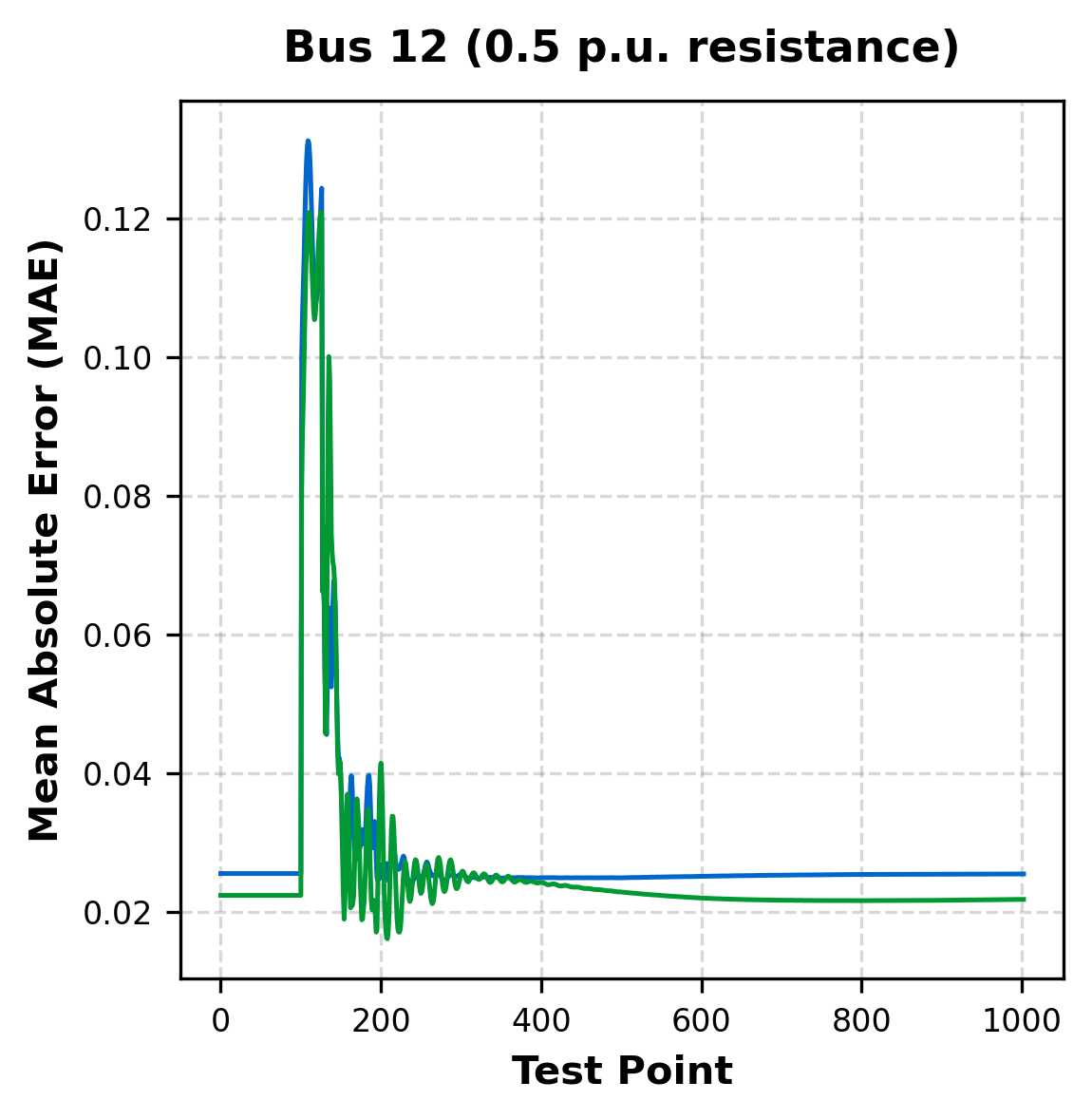}
   \caption{\textit{Scenario 4} - Testing datasets for 3-phase fault, with different per-unit resistance and locations. Graphs represent the testing set, plus \textit{Scenario} \textit{S4.1} to \textit{S4.5} from left to right.}
   \label{fig:3phase}
   \vspace{-1\baselineskip}
\end{figure*}

\begin{figure*}[t]
   \centering
   \includegraphics[width=0.0175\textwidth]{figures/labels_tilted.png}
   \includegraphics[width=0.235\textwidth]{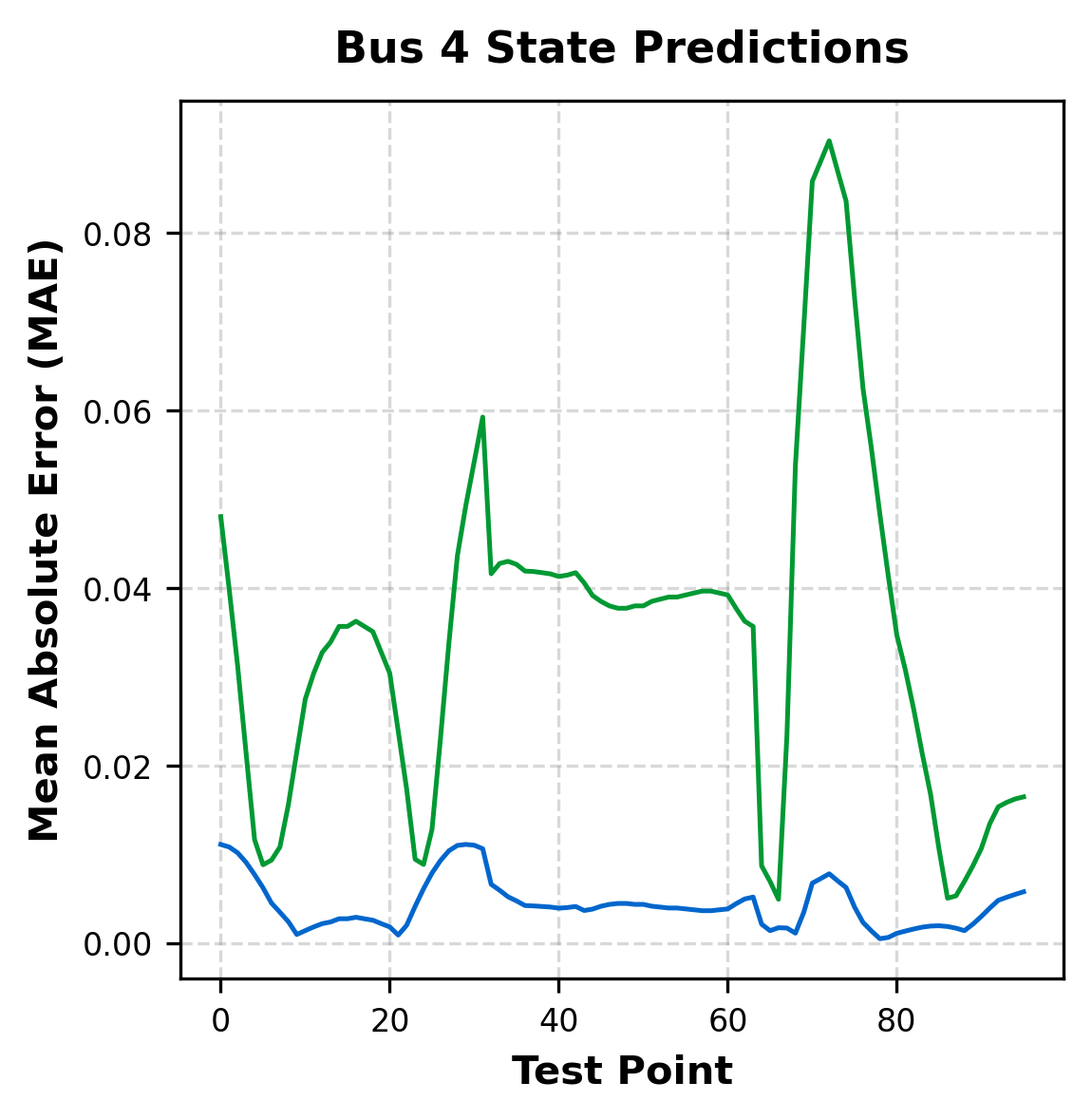}
   \includegraphics[width=0.235\textwidth]{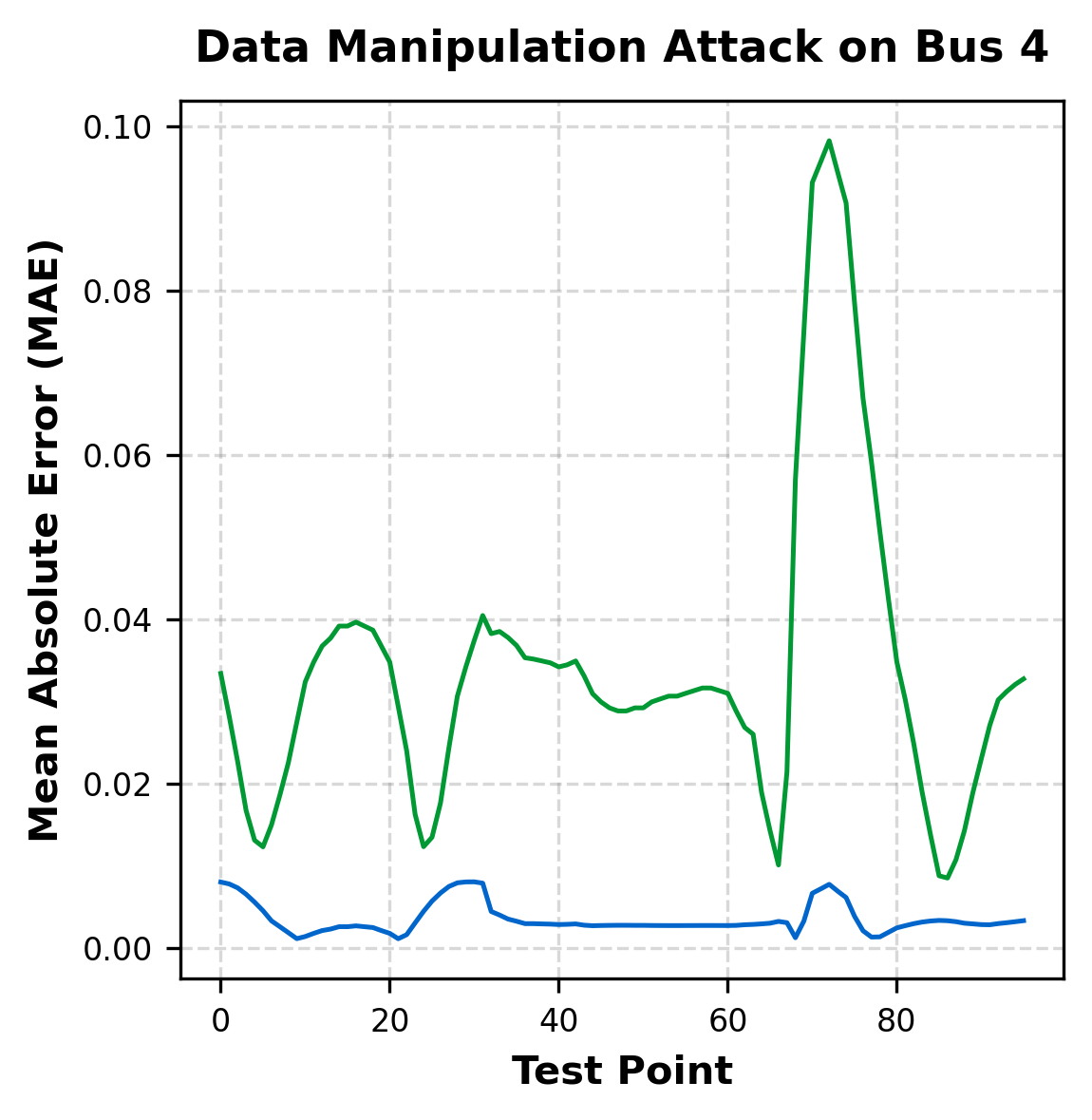}
   \includegraphics[width=0.235\textwidth]{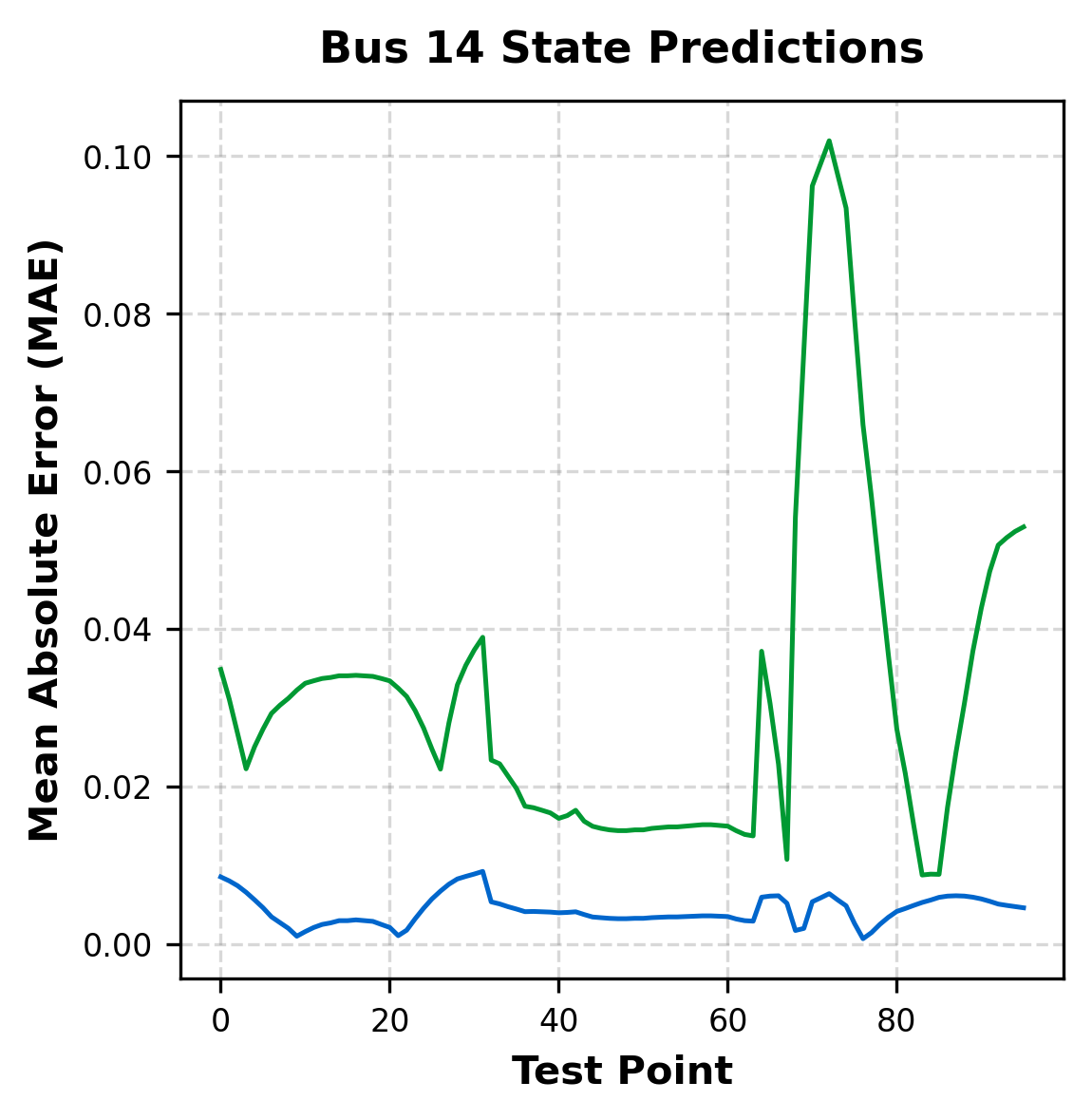}
   \includegraphics[width=0.235\textwidth]{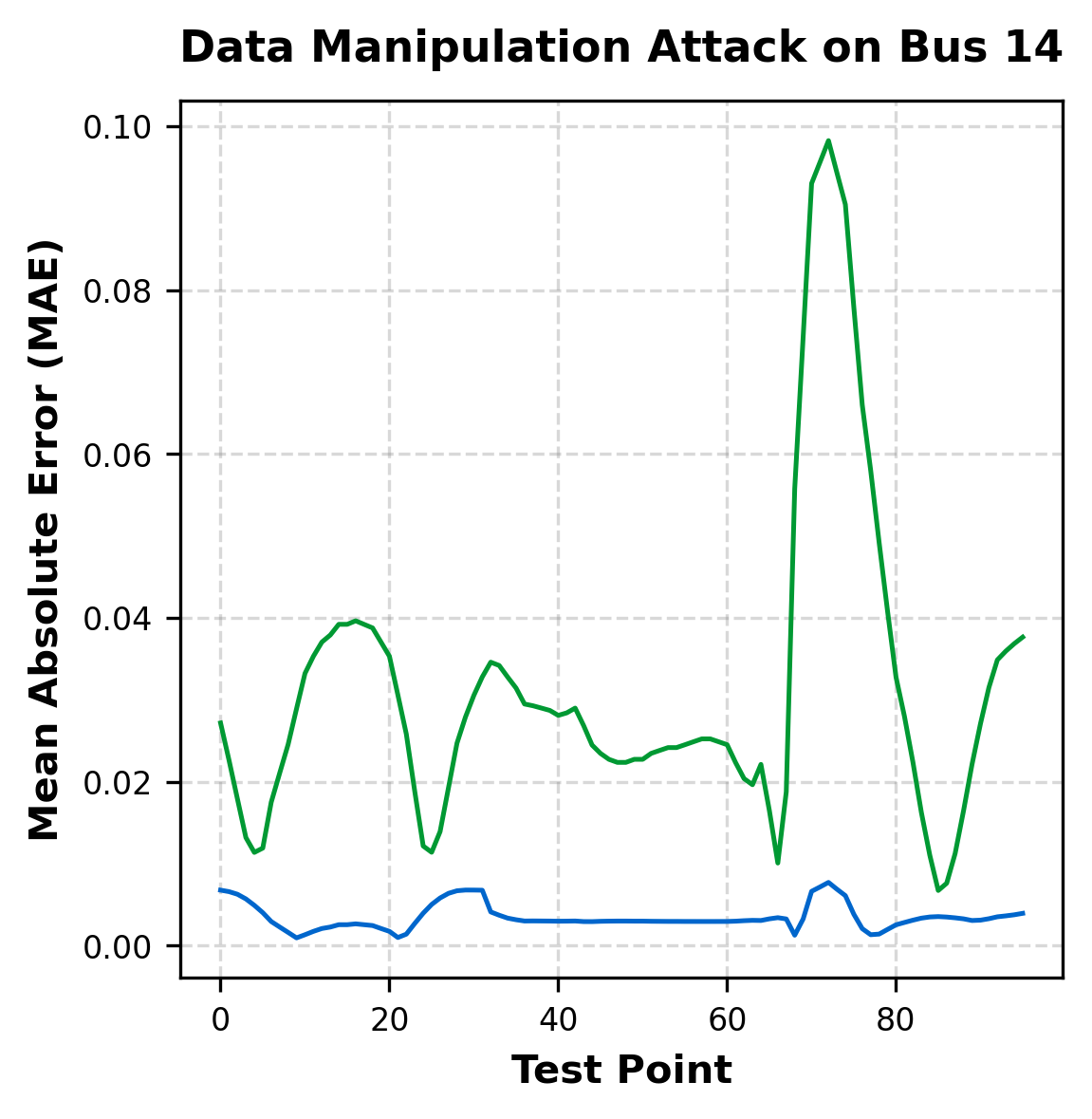}
   \caption{\textit{Scenario 5} - Testing datasets, showing the local and global accuracy of the models during Data Manipulation Attacks.}
   \label{fig:fdi_attack}
   \vspace{-\baselineskip}
\end{figure*}

\begin{table}[t]
    \centering
    \caption{Computational Costs.}
    \label{tab:computational_costs}
    \resizebox{\columnwidth}{!}{
        \begin{tabular}{||c||c|c||}
            \hline\hline
            Power System & Training Time (s) & Inference Time (ms) \\
            \hline
            IEEE 14 & 34.375 & 0.03946\\
            \hline
            IEEE 118 & 39.709 & 0.03479\\            
            \hline\hline
        \end{tabular}    }
    \vspace{-2\baselineskip}
\end{table}

\begin{table}[t]
    \centering
    \caption{Experimental Setup Scenarios (IEEE 14-bus System).}
    \label{tab:exp_scenarios}
    \resizebox{\columnwidth}{!}{
        \begin{tabular}{||c||c|c||}
            \multicolumn{3}{c}{\textbf{Scenario 1: Steady State}} \\
            \hline\hline
            \textbf{Index} & \textbf{Training Dataset} & \textbf{Testing Dataset} \\
            \hline
            S1.1 & Daily Load Profile & Unknown subset of training set\\
            \hline\hline
            \multicolumn{3}{c}{\textbf{Scenario 2: Generator Shutdown at bus 2}} \\
            \hline\hline
            \textbf{Index} & \textbf{Training Dataset} & \textbf{Testing Dataset} \\
            \hline
            S2.1 & Generator 2 Shutdown & System Load Increase 40 MW ($\sim+9\%$) \\
            \hline
            S2.2 & Generator 2 Shutdown & System Load Increase 80 MW ($\sim+18\%$)\\
            \hline\hline
            \multicolumn{3}{c}{\textbf{Scenario 3: Generation Ramp-up at bus 2}} \\
            \hline\hline
            \textbf{Index} & \textbf{Training Dataset} & \textbf{Testing Dataset} \\
            \hline
            S3.1 & Ramp-up 40 MW & System Load Decrease -40 MW ($\sim-9\%$)\\
            \hline
            S3.2 & Ramp-up 40 MW & System Load Decrease -80 MW ($\sim-18\%$)\\
            \hline\hline
            \multicolumn{3}{c}{\textbf{Scenario 4: 3-phase Transient Faults}} \\
            \hline\hline
            \textbf{Index} & \textbf{Training Dataset} & \textbf{Testing Dataset} \\
            \hline
            S4.1 & Bus 2 (0.5 p.u. resistance) & Bus 5 (0.5 p.u. resistance) \\
            \hline
            S4.2 & Bus 2 (0.5 p.u. resistance) & Bus 7 (0.5 p.u. resistance) \\
            \hline
            S4.3 & Bus 2 (0.5 p.u. resistance) & Bus 12 (0.5 p.u. resistance) \\
            \hline\hline
            \multicolumn{3}{c}{\textbf{Scenario 5: Data Manipulation Attack}} \\
            \hline\hline
            \textbf{Index} & \textbf{Training Dataset} & \textbf{Testing Dataset} \\
            \hline
            S5.1 & Daily Load Profile & Attack on bus 4 \\
            \hline
            S5.2 & Daily Load Profile & Attack on bus 14 \\
            \hline\hline
        \end{tabular}
    }
    \vspace{-2\baselineskip}
\end{table}

\subsection{Test cases and results}\label{ssec:test_cases}
A set of datasets representing different states of the IEEE 14-bus and IEEE 118-bus system benchmark are generated to evaluate the performance of the PINN approach compared to an equivalent NN in terms of accuracy and robustness in estimating power system states. The scenarios include both normal daily load profiles and transient operating conditions, to better showcase the PINN's performance under different scenarios.

\subsubsection{IEEE 14-bus System}\label{sssec:ieee_14}
To evaluate the PINN's generalization capabilities, we employ conventional dataset splitting and test the model against entirely unseen scenarios that maintain similar system dynamics. As given in Table~\ref{tab:exp_scenarios}, the testing scenarios for the IEEE 14-bus system encompass five distinct cases: (1) steady-state conditions with typical daily load profiles; (2) generator shutdown at bus 2, tested against system-wide load increases; (3) generation power surge at bus 2, validated against load reductions; (4) transient three-phase self-clearing faults (0.5 seconds duration), where the model trained on bus 2 is tested against faults at various locations (bus 5, 7, and 12); and (5) a data manipulation attack scenario, where an attacker gains control of one bus (bus 4 or 14) and tries to influence the state estimation procedure by altering sensor data feeds. This comprehensive approach assesses the model's ability to extrapolate its learned knowledge to unexpected, yet physically consistent scenarios.

\sloppypar{
In \textit{Scenario 1}, we explored combinations of weights ($\lambda_{d},\lambda_{p},\lambda_{c}$) (see Eq.~\ref{eq:loss_function}) for the data, physics, and constants terms in the loss function, using $k=3$, step size $\Delta$=0.1 and number of trials $t=10$, while maintaining $\lambda_d + \lambda_p + \lambda_c = 1$. The impact of different weight distributions on model performance is shown in Fig.~\ref{fig:steady_state}. The green curve represents the combination of weights that corresponds to a purely data-driven approach that is an optimized NN, but with no physics information, e.g. $(\lambda_{d},\lambda_{p},\lambda_{c}) = (1.0, 0.0, 0.0)$. Every combination below the green curve, shown in blue, represents different weight values among the data, physics, and constants terms.
Our results in Fig~\ref{fig:steady_state_mae_per_epoch} indicate that, in the steady-state scenario, the specific weighting for physics or constants does not significantly impact accuracy as long as the approach is not purely data-driven, each case corresponding to a different PINN with loss function as in Eq.~\ref{eq:loss_function}. The red curves above correspond to combinations that rely heavily on constants (topmost curve) or are primarily physics-based (second set of red curves) while $\lambda_{d}$ is zero.
As shown in Fig.~\ref{fig:weights_heatmap}, a strong reliance on known system constants does not lead to satisfactory results. Overall, the PINNs exhibit faster convergence and improved model accuracy, achieving lower MAE values earlier than the NN. Notably, under steady-state conditions for the IEEE 14-bus system, the PINN achieves up to an $\sim57.3\%$ reduction in MAE, demonstrating its superior performance.
}

\textit{Scenario 2} test results can be seen in Fig.~\ref{fig:shutdown}. In the initial testing phase, the PINN does not offer substantial advantages over the NN. However, looking at \textit{Scenario} \textit{S2.1} and \textit{S2.2}, we can see that the PINN offers significantly improved accuracy and stability in the predictions. While the NN's predictions seem to deteriorate over time, the PINN has a lot more stable MAE over time. In terms of accuracy, the PINN shows $\sim23.5\%$ and $\sim8\%$ improvements in absolute error for \textit{Scenario} \textit{S2.1} and \textit{S2.2}, respectively, indicating the better generalization of the model to unknown states of the system.

Fig.~\ref{fig:increase} shows the testing results for \textit{Scenario 3}. The NN's prediction accuracy seems to vary significantly in MAE based on the data points and shows significantly more fluctuations than the PINN. The PINN, on the other hand, stays consistently low in MAE across all data points while also being more accurate. Examining the unknown testing scenarios, the PINN consistently stays more accurate than the NN in all data points, again, hinting the PINN's greater generalization potential to unknown system states. Specifically, the PINN has $\sim65.15\%$ in \textit{S3.1} and $\sim33.23\%$ in \textit{S3.2}, lower MAE than the NN.

Regarding \textit{Scenario 4}, as depicted in Fig.~\ref{fig:3phase}, we evaluate the model's ability to extrapolate fault characteristics across varying locations. The proposed PINN demonstrates, on average, $\sim20.53\%$ better accuracy when compared to the NN, even though both models underwent identical hyperparameter tuning.
The impact of fault location on model performance is detailed in Fig.~\ref{fig:3phase}'s multiple graphs, showing the testing accuracy along with the different scenarios (from \textit{S4.1} to \textit{S4.3}, from left to right). While PINNs exhibit strong generalization capabilities for faults occurring near training locations, their accuracy diminishes as the spatial distance between training and testing locations increases, underscoring the significance of spatial correlation.

In data manipulation attacks, malicious actors strategically manipulate sensor data to undermine the integrity of real-time grid monitoring. To investigate the robustness of the proposed PINN in such attacks, we tested the PINN's accuracy, which can act as a filter against such erroneous data. Such filtering properties enhance continuous and accurate state monitoring capabilities, even under data communication disruptions. To evaluate the robustness of the IEEE 14-bus system's state estimation against data manipulation attacks, we target buses 4 and 14 in \textit{Scenario 5} (Table~\ref{tab:exp_scenarios}).

Models, including both PINN and the NN, are trained using the steady-state dataset described in \textit{S1.1}. To evaluate the PINN and NN performance, data manipulation is applied to the testing dataset. This process is employed to systematically evaluate the performance of the models under varying levels of data corruption, enabling the assessment of their tolerance to increasing measurement biases. The parameters governing this data manipulation process are defined as follows: Attack targets are chosen, namely bus 4 and bus 14. The buses were selected based on the number of adjacent buses to the target bus, with \textit{S5.1} using bus 4 due to having the most adjacent buses and \textit{S5.2} using bus 14 for having the least.
The attack specifically targets active and reactive power injection measurements ($P$ and $Q$) at the selected bus. Across 100 consecutive time instances, the bias magnitude profile is incrementally increased relative to the true measurement value at each test point: 10\% for the first third (test points 1-33), 20\% for the second third (test points 34-66), and 30\% for the final third (test points 67-100). This profile is applied sequentially over the 100 test instances, modifying only the targeted $P$ and $Q$ measurements of the compromised bus, while all other measurements remain unaffected. This systematic application allows for evaluating the models' response and performance degradation under progressively larger measurement errors.

As shown in Figure~\ref{fig:fdi_attack}, the PINN significantly outperforms the equivalent NN in predicting the state of the affected bus and the overall system state, demonstrating its ability to leverage underlying physical laws to filter out the effects of the data manipulation attack. The PINN achieves up to $\sim93\%$ improvement in average MAE for both \textit{S5.1} and \textit{S5.2}, highlighting its robustness against such attacks.

\subsubsection{IEEE 118-bus System}\label{sssec:IEEE_118}
To demonstrate the scalability of the approach, experimental scenarios are applied to the IEEE 118-bus system, similar to the test cases seen in Section~\ref{sssec:ieee_14}. Specifically, the testing scenarios are: (1) steady-state conditions with typical daily load profiles; (2) a generator shutdown scenario at bus 26; (3) transient three-phase self-clearing fault at bus 19 (0.5 seconds duration); and (4) a data manipulation attack scenario, where an attacker gains control of bus 54 and alters the power injection measurements. The specifics of each scenario can be seen in Table~\ref{tab:exp_scenarios_118}.

\begin{table}[t]
    \centering
    \caption{Experimental Setup Scenarios (IEEE 118-bus System).}
    \label{tab:exp_scenarios_118}
    \resizebox{\columnwidth}{!}{
        \begin{tabular}{||c||c|c||}
            \multicolumn{3}{c}{\textbf{Scenario 6: Steady State}} \\
            \hline\hline
            \textbf{Index} & \textbf{Training Dataset} & \textbf{Testing Dataset} \\
            \hline
            S6.1 & Daily Load Profile & Unknown subset of training set\\
            \hline\hline
            \multicolumn{3}{c}{\textbf{Scenario 7: Generator Shutdown at bus 26}} \\
            \hline\hline
            \textbf{Index} & \textbf{Training Dataset} & \textbf{Testing Dataset} \\
            \hline
            S7.1 & Generator 26 Shutdown (-314MW) & System Load Increase 314 MW \\
            \hline\hline
            \multicolumn{3}{c}{\textbf{Scenario 8: 3-phase Transient Faults}} \\
            \hline\hline
            \textbf{Index} & \textbf{Training Dataset} & \textbf{Testing Dataset} \\
            \hline
            S8.1 & Bus 19 (0.25 p.u. resistance) & Bus 33 (0.25 p.u. resistance) \\
            \hline\hline
            \multicolumn{3}{c}{\textbf{Scenario 9: Data Manipulation Attack}} \\
            \hline\hline
            \textbf{Index} & \textbf{Training Dataset} & \textbf{Testing Dataset} \\
            \hline
            S9.1 & Daily Load Profile & Attack on bus 54 \\
            \hline\hline
        \end{tabular}
    }
    \vspace{-2\baselineskip}
\end{table}

\begin{figure}[t]
    \centering
    \subfloat[MAE per training epoch\label{fig:steady_state_118_mae_per_epoch}]{
        \includegraphics[width=0.295\textwidth]{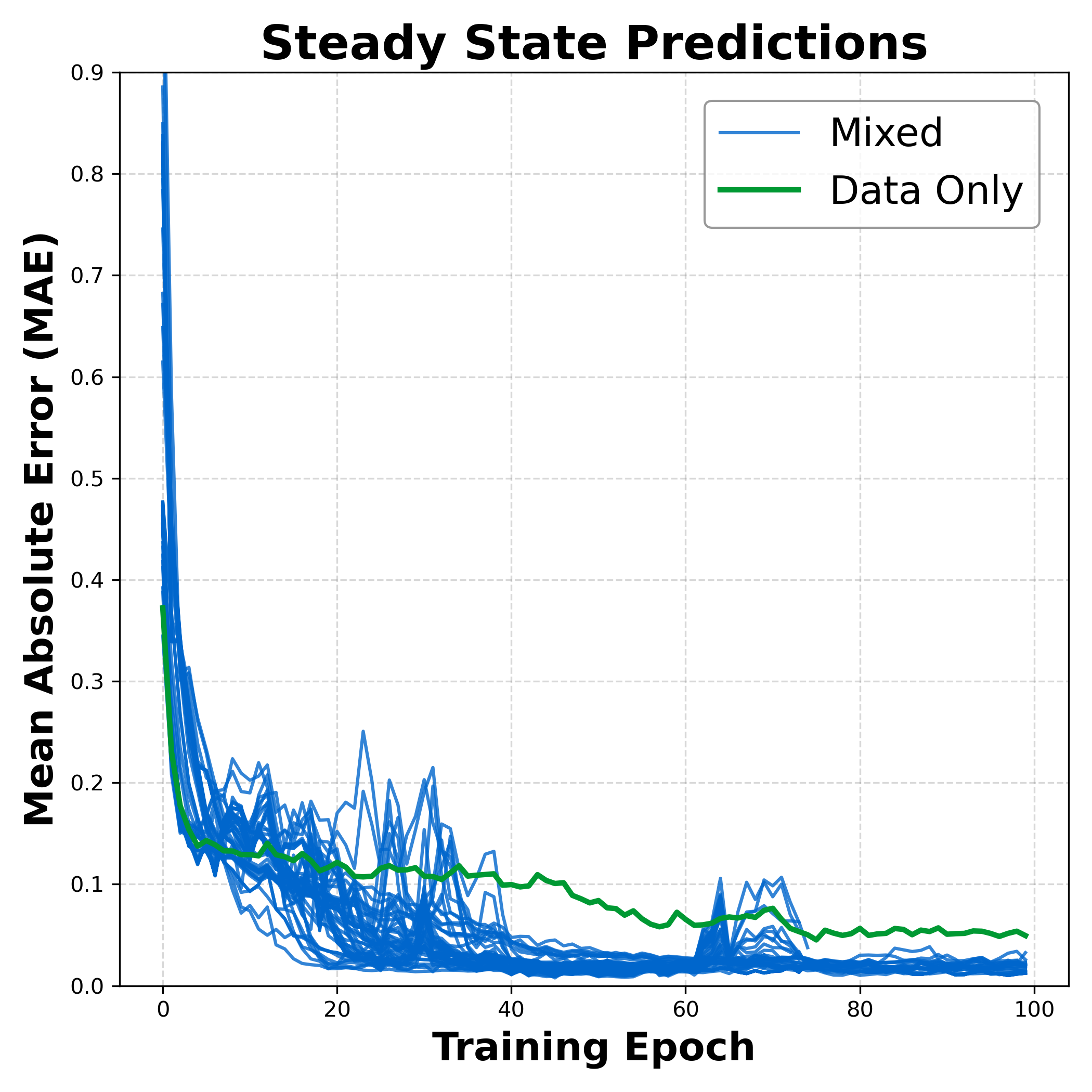}
    }
    \hfill
    \subfloat[$\lambda$ weights heatmap\label{fig:weights_heatmap_118}]{
        \includegraphics[width=0.395\textwidth]{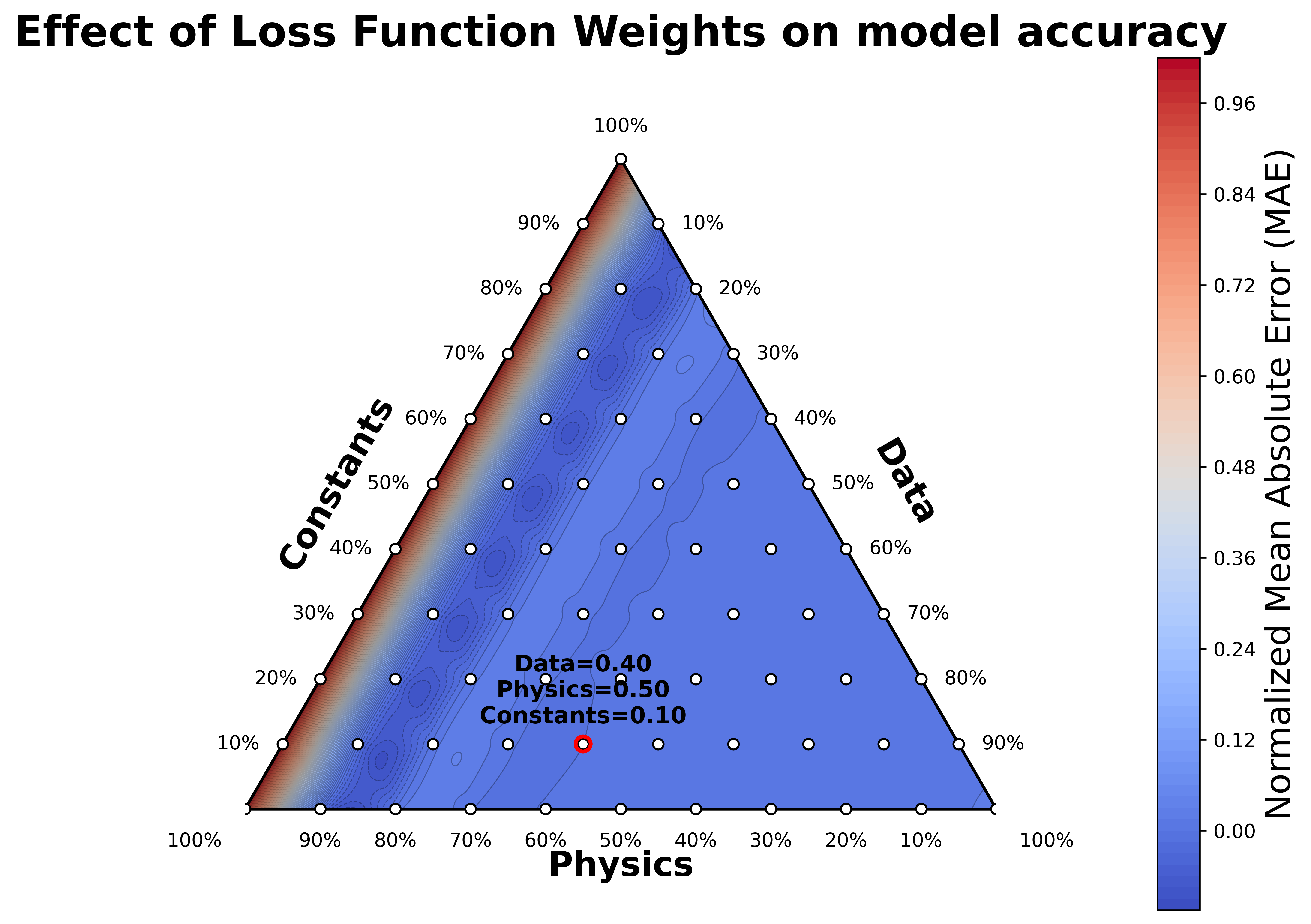}
    }
    \caption{\textit{Scenario 6 (IEEE 118-bus System)} - MAE for steady-state conditions using varying $\lambda_{d},\lambda_{p},\lambda_{c}$ weights for data, physics, and constants, respectively, in the loss function.}
    \label{fig:steady_state_118}
    \vspace{-2em}
\end{figure}

In \textit{Scenario 6}, we follow the same principles as in \textit{Scenario 1}, a steady-state daily load profile adapted to the larger system. In particular, the goal is to examine the effect of the different weights of the loss function parameters on a model trained using the power system's steady state conditions.
Fig.~\ref{fig:steady_state_118} shows the different PINN configurations in blue and the benchmark NN in green. The red "No Data" curves, as seen in Fig.~\ref{fig:steady_state} are omitted here due to their unsatisfactory performance, lying outside the region of interest shown in this graph.
While the PINNs show a more erratic behavior during early training epochs than in the smaller IEEE benchmark system, the end result confirms the superiority of the PINN over the NN. In particular, the final results show even greater gains in accuracy, with the best PINN configuration achieving $\sim82.91\%$ lower MAE than the NN. The loss function weight heatmap (Fig.~\ref{fig:weights_heatmap_118}) highlights similar regions of interest as the IEEE 14-bus system.

\begin{figure}[t]
    \centering
    \includegraphics[width=0.0175\textwidth]{figures/labels_tilted.png}
    \includegraphics[width=0.22\textwidth]{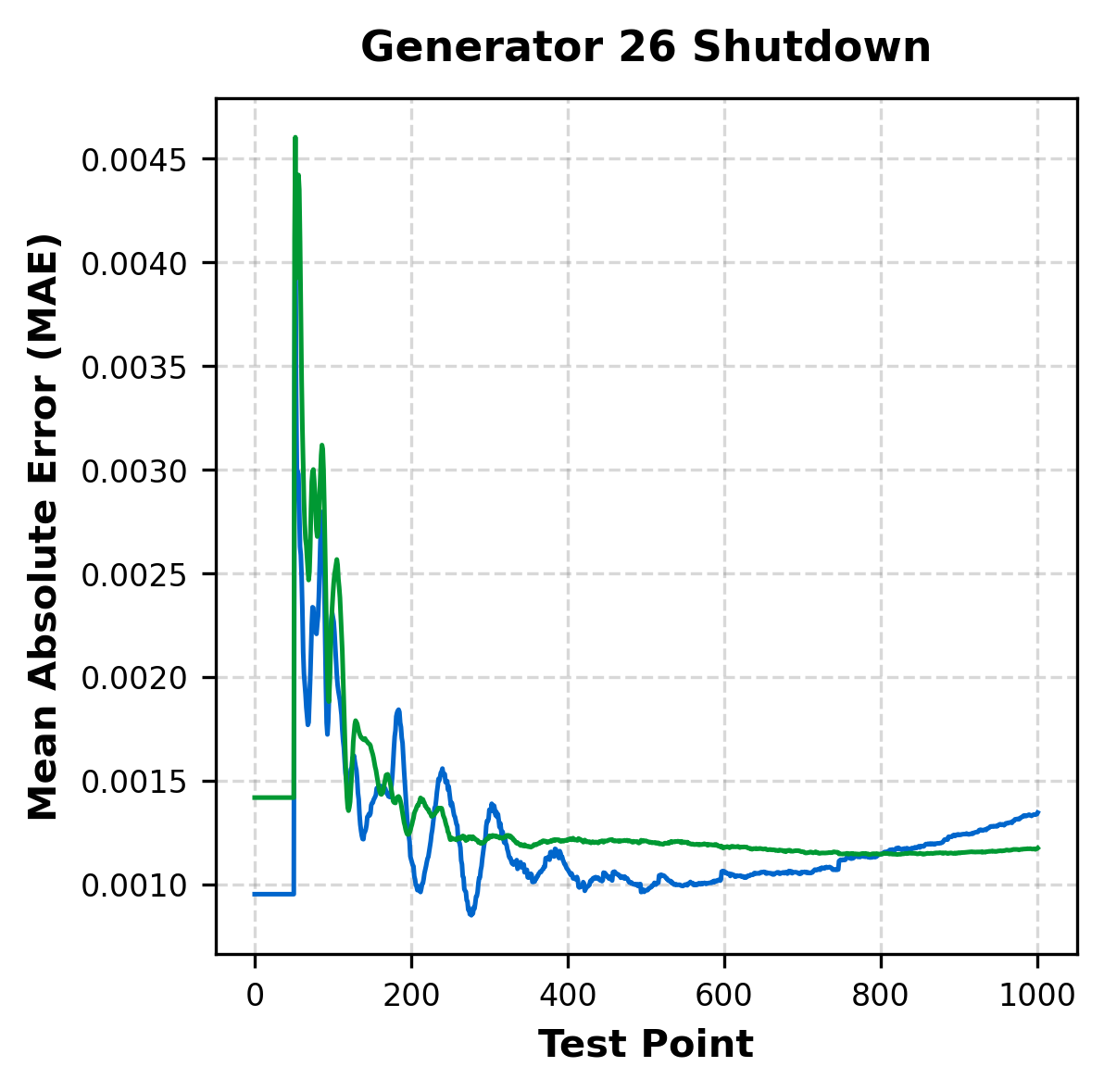}
    \includegraphics[width=0.22\textwidth]{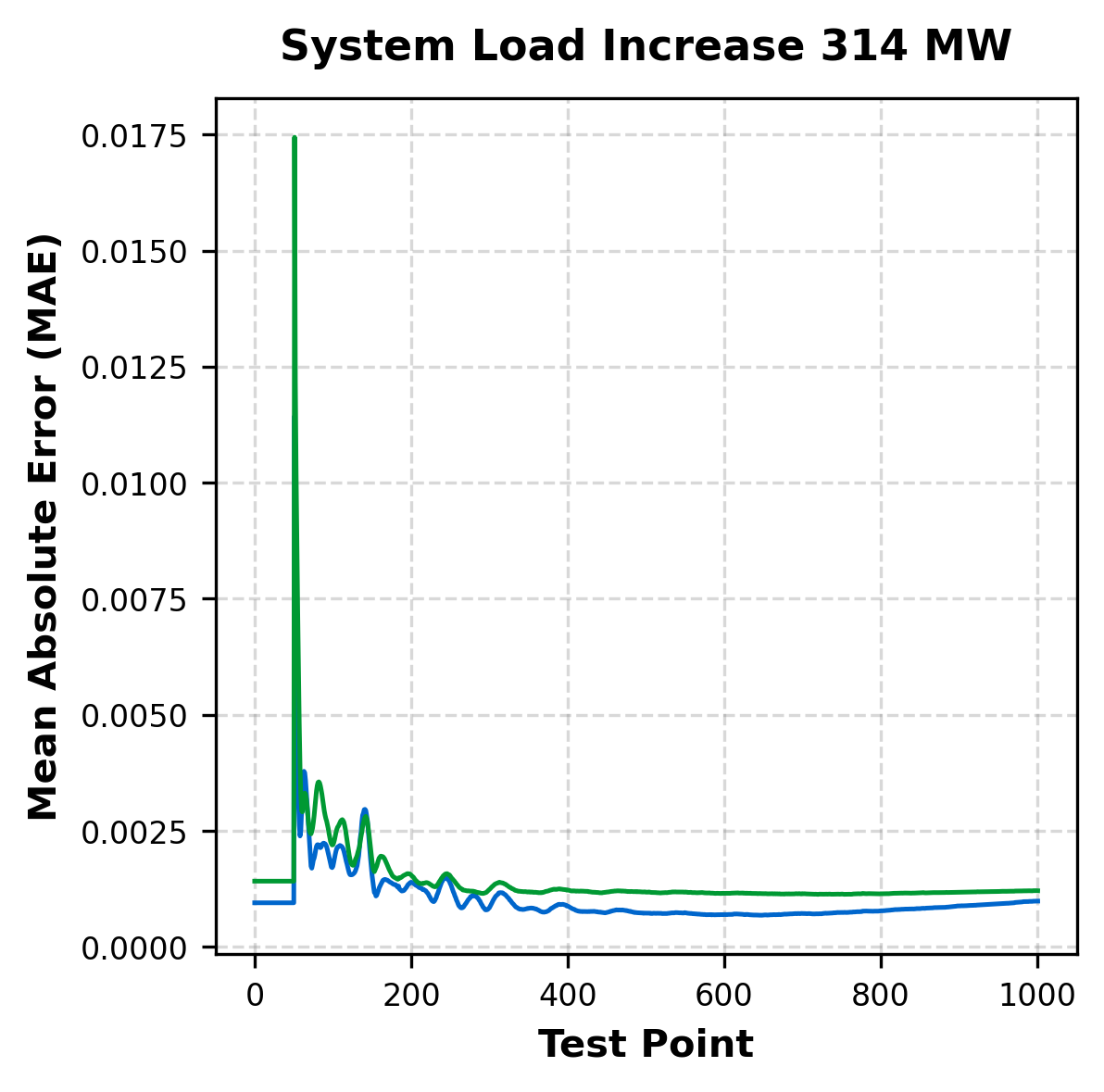}
    \caption{\textit{Scenario 7} - Testing datasets with generator shutdown. Models trained on a bus 26 shutdown are tested with load spikes at equal (\textit{S7.1}) the initial shutdown magnitude.}
    \label{fig:shutdown_118}
    \vspace{-1em}
\end{figure}

\textit{Scenario 7} examines a generator shutdown. The model is trained using a dataset capturing the abrupt interruption of the generator at bus 26 and tested against a dataset representing an equivalent system load increase of 314 MW, distributed equally across buses 20 to 23 and bus 27. Overall, the PINN shows less extreme behavior on the initial testing dataset, while the NN seems to stabilize faster, as seen in Fig.~\ref{fig:shutdown_118}. The PINN's robustness capabilities show up when testing against the equivalent load increase scenario, where it shows a clear trend, showing a $~27.68\%$ lower MAE.

\begin{figure}[t]
    \centering
    \includegraphics[width=0.0175\textwidth]{figures/labels_tilted.png}
    \includegraphics[width=0.22\textwidth]{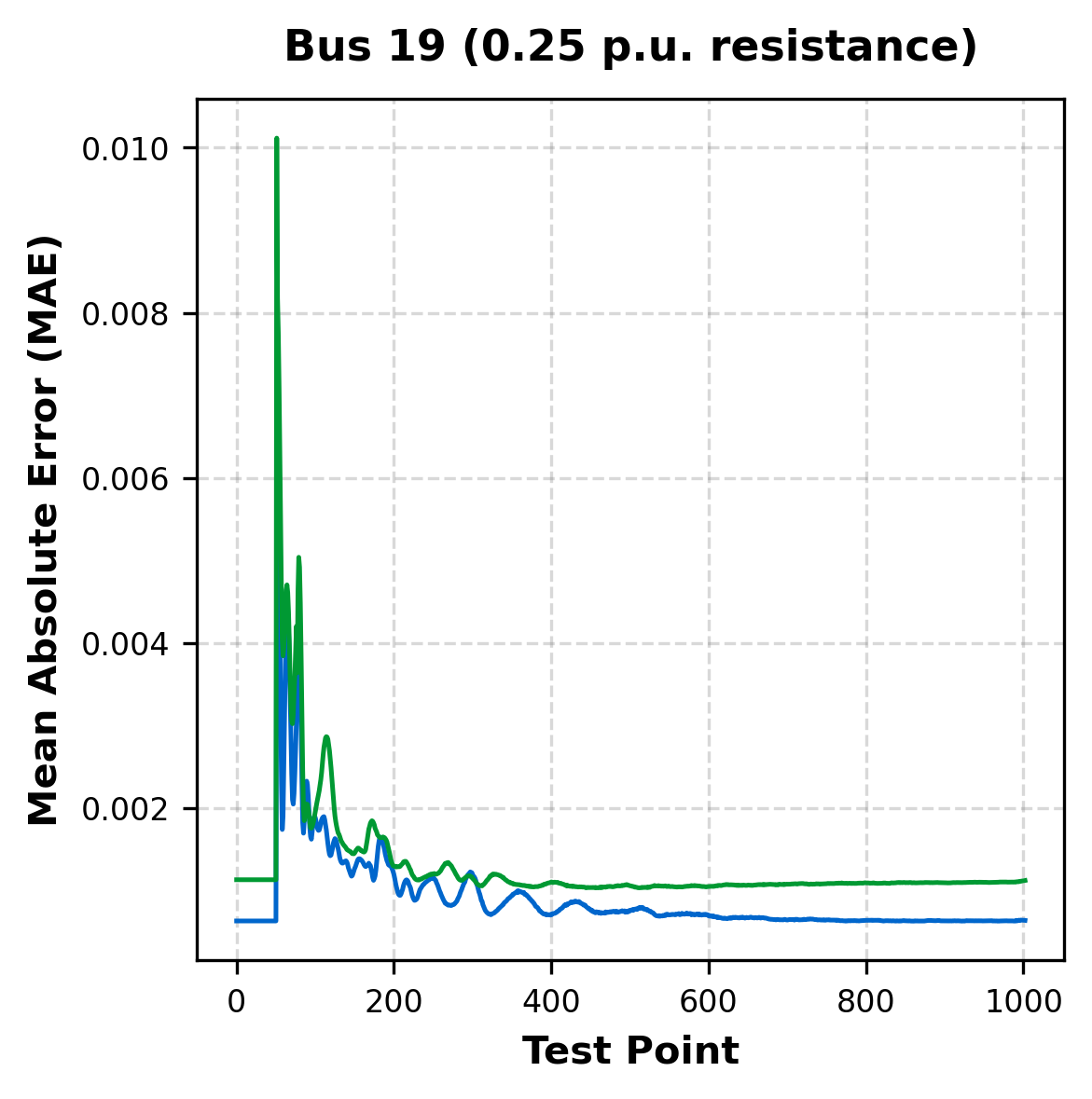}
    \includegraphics[width=0.22\textwidth]{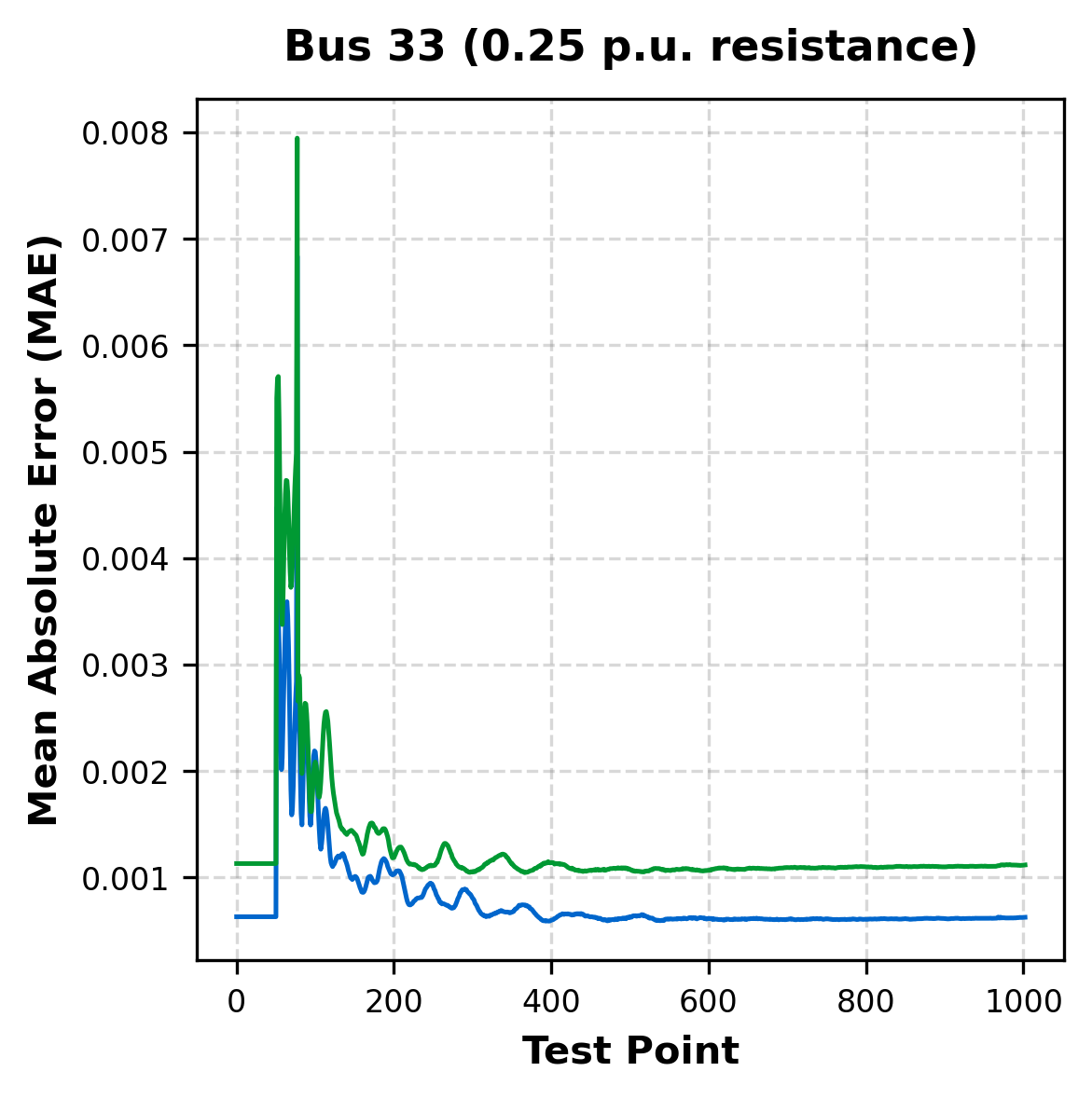}
    \caption{\textit{Scenario 8} - Testing datasets for 3-phase fault. Graphs represent the testing set, plus \textit{Scenario} \textit{S8.1}}
    \label{fig:3phase_118}
    \vspace{-2em}
\end{figure}

Fig.~\ref{fig:3phase_118} shows the testing results for the 3-phase transient fault, namely \textit{Scenario 8}. In this scenario, the models are trained with a transient 3-phase fault at bus 19 and tested against measurements representing a same-type fault in a different location, bus 33. The PINN generates more accurate results when tested against both the unknown subset of the training dataset and the completely unknown dataset, with a $\sim37.67\%$ improvement on the testing dataset.

\begin{figure}[t]
    \centering
    \includegraphics[width=0.0175\textwidth]{figures/labels_tilted.png}
    \includegraphics[width=0.22\textwidth]{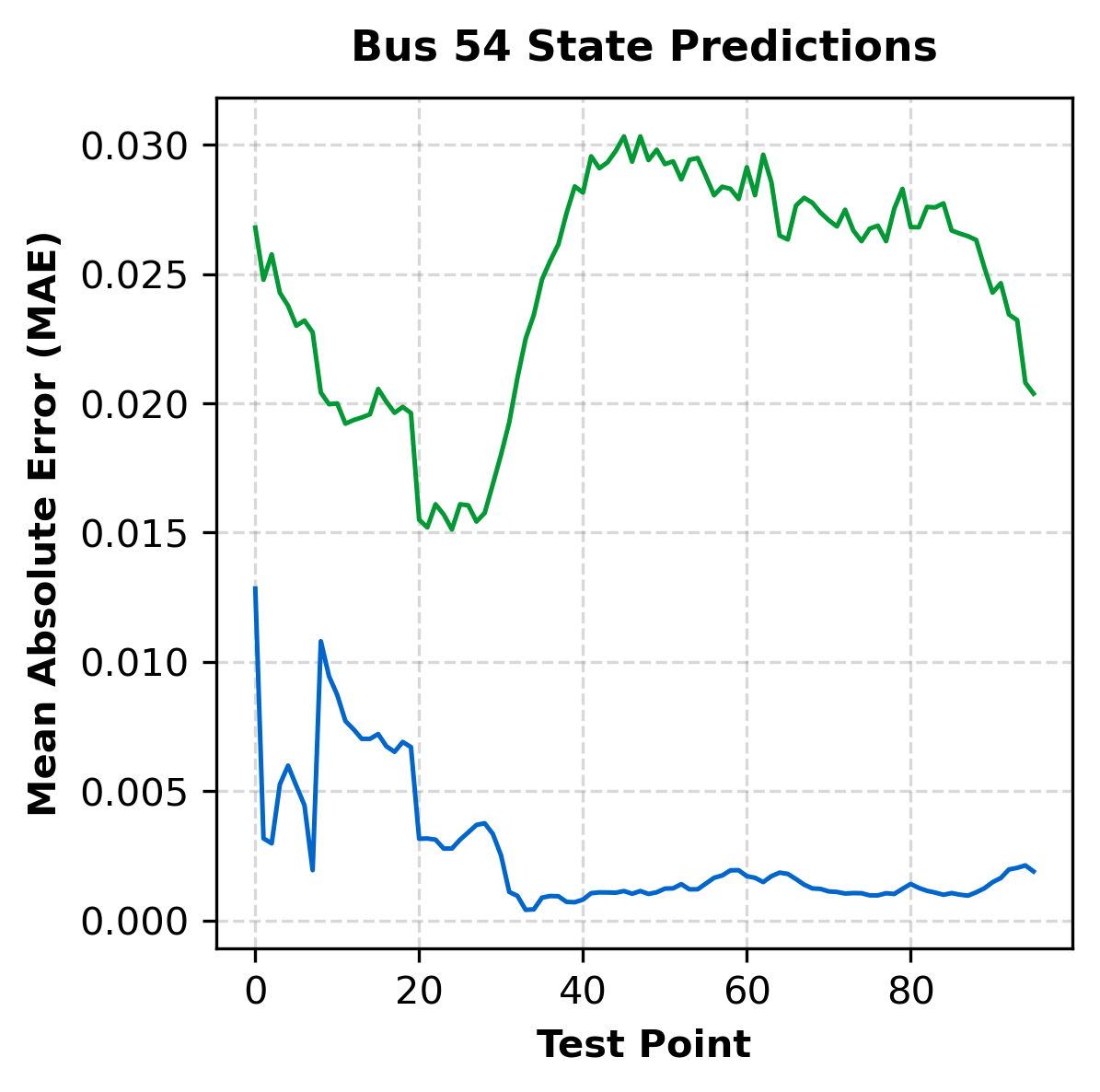}
    \includegraphics[width=0.22\textwidth]{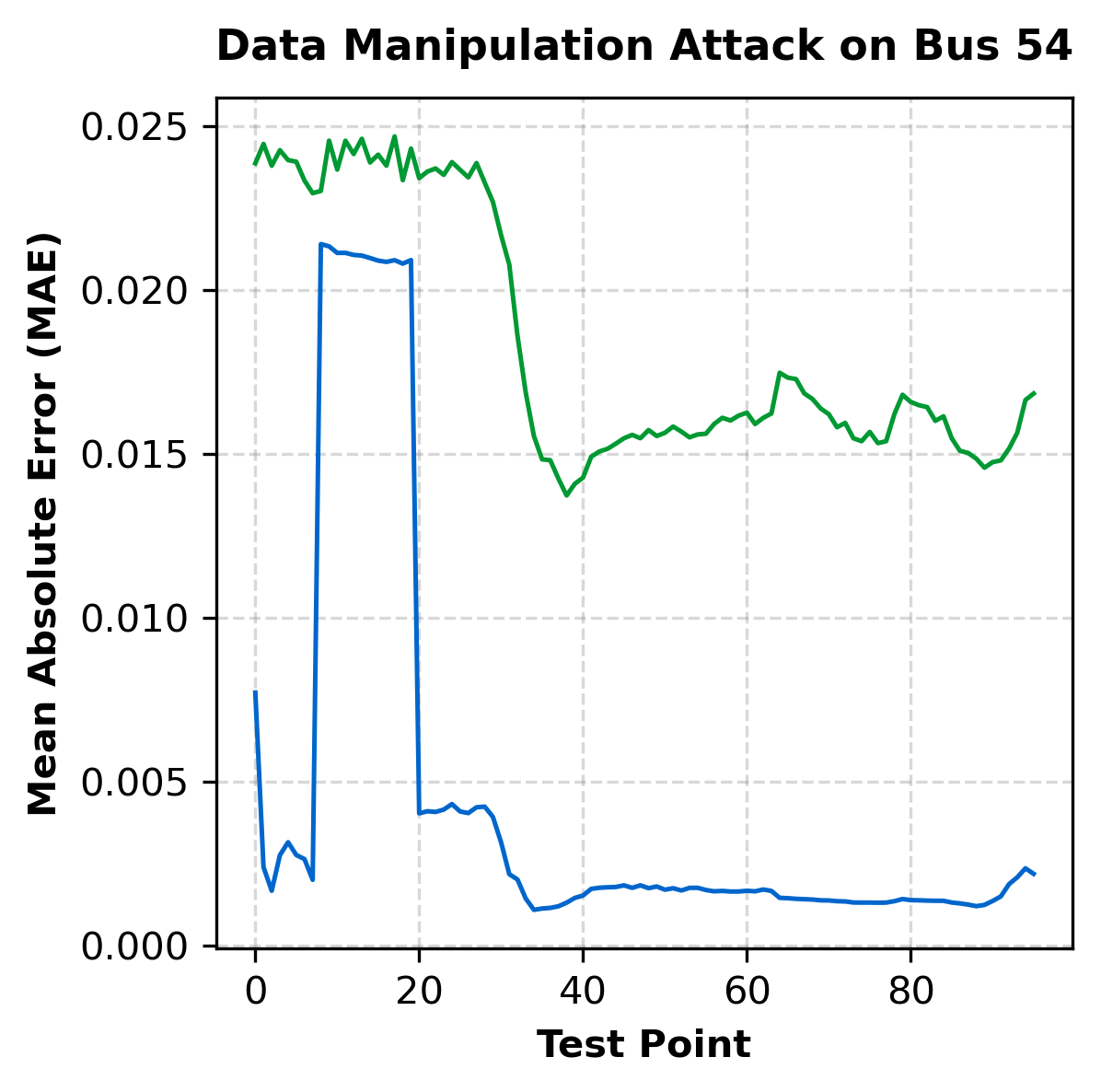}
    \caption{\textit{Scenario 9} - Testing datasets showing the local and global accuracy of the models during Data Manipulation Attacks.}
    \label{fig:fdi_attack_118}
    \vspace{-1em}
\end{figure}

Finally, \textit{Scenario 9}, as depicted in Fig.~\ref{fig:fdi_attack_118}, follows the same training and testing principles outlined for \textit{Scenario 5} in \ref{sssec:ieee_14}. The figure shows the overall average MAE of the system's predictions as well as the MAE specifically for bus 54, the attack target. Bus 54 was selected because it has a high load of 113 MW, making it one of the largest buses in the system and highly influential to the entire network.
The models, both PINN and NN, are trained using the same dataset as in \textit{Scenario 6} and tested against a dataset that represents the increasing data manipulation magnitude on bus 54. The PINN shows considerable improvements of up to $\sim76.11\%$ in accurately predicting the system-wide state.

\subsubsection{Experimental Results Discussion}\label{sssec:exp_discussion}
The PINN has been comprehensively tested in a large variety of datasets and has shown adequate robustness in terms of model inference. State estimation results demonstrate superior accuracy, when compared to an equivalent NN, across the 14-bus and 118-bus IEEE benchmark systems, highlighting the scalability of the approach. The associated heatmaps for the steady state scenarios show that the majority of good solutions lie in the space of $\lambda_d \simeq 10\%-20\%$, with different combinations of physics and data showing similar results. However, this does not discount the possibility of finding the optimal solution elsewhere (see Fig.~\ref{fig:steady_state_118}), but it can serve as an indication on how to reduce the hyperparameter tuning search space, if needed.

Table~\ref{tab:exp_results_comparison} presents a comparative summary of selected methods from the literature applied to the IEEE 118-bus system, based on the accuracy reported by the respective authors. To facilitate relative evaluation, all values are normalized with respect to the performance of the proposed PINN, which is set to 1.0. The comparison draws exclusively from results reported in the original publications, without re-implementation or empirical replication. Due to differences in datasets, problem formulations, and model architectures, the table is intended as a representative summary rather than a strict benchmark. Under the reported conditions, the normalized metrics indicate that the proposed PINN achieves superior performance relative to existing methods.

\begin{table}[t]
    \centering
    \caption{State Estimation Performance (IEEE 118-bus System).}
    \label{tab:exp_results_comparison}
    \resizebox{\columnwidth}{!}{
        \begin{tabular}{||c||c||c||}
            \hline\hline
            Method & Mean Absolute Error & Normalized Performance\\
            \hline\hline
            PINN & $7.72 \cdot 10^{-3}$ & 1.00\\
            \hline
            GNU-GNN~\cite{yang2022data} & $9.42 \cdot 10^{-3}$ & 0.82\\
            \hline
            Prox-linear Net~\cite{zhang2019unrolled} & $1.26 \cdot 10^{-2}$ & 0.61\\
            \hline
            NN & $4.52 \cdot 10^{-2}$ & 0.17\\
            \hline\hline
        \end{tabular}
    }
    \vspace{-2em}
\end{table}

\section{Conclusion}\label{sec:conclusion}
This work contributes to the growing body of research on integrating data-driven and physics-based methodologies, demonstrating the potential of PINNs to address the dynamic challenges of modern power systems. By embedding physical laws directly into the neural network architecture and systematically optimizing the model and hyperparameters, PINNs enhance the accuracy of power system state estimation, outperforming traditional machine learning models, particularly during disruptions. Importantly, our experiments show that PINNs are less prone to data manipulation attacks, a common threat to power systems, as they leverage the underlying physical constraints rather than relying solely on observed data, highlighting this approach's robustness compared to traditional techniques.

Future work could explore the adoption of a multi-PINN architecture, where each model is trained on localized data from specific regions or clusters within the power system. Clustering based on proximity or electrical characteristics could optimize the models' ability to capture local dynamics and fault behaviors. Additionally, leveraging transfer learning between adjacent clusters may enhance inter-cluster fault prediction accuracy, thereby improving overall system reliability. Extending this framework with a physics-aware implementation opens an intriguing avenue for anomaly detection. By identifying deviations from expected system behavior, this approach could enable advanced monitoring and early detection of anomalies, providing valuable insights even in complex scenarios or under limited system observability.
\vspace{-1em}

\bibliographystyle{IEEEtran}
\bibliography{sources}

\end{document}